\documentclass[3p]{elsarticle}
\usepackage{xpatch}

\usepackage[utf8]{inputenc}
\usepackage[english]{babel}
\usepackage[table]{xcolor} 
\usepackage{graphicx} 
\usepackage{subcaption} 
\usepackage{pdfpages} 
\usepackage{amssymb} 
\usepackage{amsmath}
\usepackage{tabularx} 
\usepackage{booktabs} 
\usepackage{longtable}
\usepackage{multirow} 
\usepackage{makecell} 
\usepackage{array}
\usepackage[most]{tcolorbox} 
\usepackage{tikz}
\usepackage{threeparttable} 
\usetikzlibrary{fit}
\usepackage{soul}

\input{glyphtounicode}
\usepackage{longtable}

\usepackage[pagebackref=true,breaklinks=true]{hyperref}
\renewcommand*\backref[1]{\ifx#1\relax \else (\^ #1) \fi}
\hypersetup{
    colorlinks,
    linkcolor={blue},
    citecolor={blue},
    urlcolor={blue}
}
\DeclareMathOperator*{\argmax}{argmax}

\usepackage{cleveref}

\usepackage{color}

\title{How to Choose a Reinforcement-Learning Algorithm}

\author{Fabian~Bongratz$^{1,2}$, Vladimir~Golkov$^{1,2}$, Lukas~Mautner$^{1}$, Luca~Della~Libera$^{1}$, \\ Frederik~Heetmeyer$^{1}$, Felix Czaja$^{3}$, Julian Rodemann$^{3}$, Daniel~Cremers$^{1,2}$ \\ \vspace{5pt} $^1$Computer Vision Group, Technical University of Munich \\ $^2$Munich Center for Machine Learning \\ $^3$Department of Statistics, Ludwig-Maximilians-Universit\"at Munich}
\date{March 2021}

\newcommand{\E}{\ensuremath{\mathbb{E}}}
\newcommand{\Tmax}{\ensuremath{t_{\mathrm{max}}}}

\newcommand{\deceptive}{Rewards are \emph{deceptive} if they could be misleading; an example for a deceptive reward function would be the distance to a goal state that may lead the agent towards some pitfall in between \cite{mazoure2020}.}

\newcommand{\tablehead}{\textbf}

\usepackage{enumitem} 
\newlist{titemize}{itemize}{2}
\setlist[titemize]{leftmargin=*, nosep,label=\textbullet}
\makeatletter
\def\ps@pprintTitle{%
 \let\@oddhead\@empty
 \let\@evenhead\@empty
 \def\@oddfoot{\centerline{\thepage}}%
 \let\@evenfoot\@oddfoot}
\makeatother

\begin{document}

\begin{abstract}
	The field of reinforcement learning offers a large variety of concepts and methods to tackle sequential decision-making problems. This variety has become so large that choosing an algorithm for a task at hand
	can be challenging. In this work, we streamline the process of choosing reinforcement-learning algorithms and action-distribution families. We provide a structured overview of existing methods and their properties, as well as guidelines for when to choose which methods. An interactive version of these guidelines is available online at \url{https://rl-picker.github.io/}.
\end{abstract}

\maketitle

\section{Introduction} \label{sec:introduction}

\emph{Reinforcement learning} (RL) has long been known as a meaningful approach to tackling problems that could be cast into a sequential decision-making process together with a reward-like performance measure. With many fundamental concepts of RL being already established for several decades \cite{bellman1954, sutton1988, watkins1989, watkins1992, rummery1994}, the strong progress in learning meaningful state representations with deep feedforward neural networks \cite{lecun2015, goodfellow2016} paved the way for high-impact machine learning systems outperforming humans in a variety of game-related tasks \cite{silver2017, schrittwieser2020, badia2020}. Modern applications of RL are game-playing~\cite{shao2019}, robotics~\cite{kober2013}, communication systems~\cite{luong2019}, and finance~\cite{charpentier2021}, to name just a few.

Nowadays, deep reinforcement learning encompasses a large number of methods and algorithms with varying properties. Unfortunately, the information about which method properties to choose in which situations is distributed across the considerable length of standard courses and textbooks; and the information about which method has which properties is distributed across dozens of papers that propose the methods.

Existing surveys about RL \cite{li2018, zhang2020, justesen2020, yu2019, kaelbing1996, weng2018, vitay2022} typically provide a compact overview of methods and application fields but lack structured guidelines for when to choose which methods. On the other hand, practical recommendations for RL algorithm design \cite{andrychowicz2021, reda2020, henderson2018, engstrom2020} mostly focus on a few environmental conditions like simulated locomotion benchmarks.   

In this work, we streamline the process of choosing RL algorithms. We achieve this by bringing the information about which algorithmic properties to choose in which situations into a quickly accessible form. In addition, we elaborate on reasons that make certain methods advantageous over others in certain environments based on practical considerations and existing literature in the field.

Unless stated otherwise, we consider a Markov decision process with a~set~$\mathcal{S}$ of possible states~$s \in \mathcal{S}$, a~set~$\mathcal{A}$ of possible actions~$a\in \mathcal{A}$, state-transition probabilities $p(s'\mid s,a) = p(s_{t+1} = s'\mid s_t = s, a_t = a)$, and rewards $r_t$, where $t \in \mathbb{N}_0$ is a timepoint, $s,s' \in \mathcal{S}$ and $a \in \mathcal{A}$. For control tasks, the goal is to learn a policy $\pi(a\mid s)$, mapping from each state~$s$ to a \emph{mixed} next action, i.e.\ a distribution over actions that aims to maximize the expectation $\mathbb{E} [ \sum_{t=0}^{\Tmax} r_{t+1} ]$ of the \emph{cumulative reward} $\sum_{t=0}^{\Tmax} r_{t+1}$ (the cumulative reward is also called \emph{return}) across an episode, where $r_t$ is the reward at time $t$ and $\Tmax$ is the duration of an episode. Value functions assign the expected future cumulative reward to either a single state (state-value function $V(s) = \mathbb{E}(\sum_{\tau=t}^{\Tmax} r_{\tau+1} \mid  s_t=s)$) or to state-action pairs (action-value function $Q(s, a) = \mathbb{E}(\sum_{\tau=t}^{\Tmax}r_{\tau+1} \mid  s_t=s, a_t=a)$ and advantage function $A(s, a) = Q(s, a) - V(s)$), assuming a fixed policy for all future decision steps. The state-value function $V(s)$ is the expectation of $Q(s,a)$ with respect to the probability measure $\pi(a\mid s)$. Note that, for decision processes with infinite time horizon, i.e.\ $\Tmax \rightarrow \infty$, the cumulative reward needs to be calculated as $\sum_{t=0}^{\Tmax} \gamma^t r_{t+1}$, where the discount factor $0 < \gamma < 1$ ensures convergence of the series. We ignore this fact for the sake of simplicity wherever possible.

The rest of this paper is structured as follows. \Cref{sec:algorithms} provides an overview of RL algorithms and guidelines for choosing them based on their key properties. \Cref{sec:policies} provides an overview of action-distribution families that can be used in the RL algorithms as a blueprint for the policy and guidelines for choosing them. Final selection steps are in \Cref{sec:final-steps}. Moreover, we provide a list of deep neural-network architectures that are popular in RL in \Cref{sec:architectures} and practical guidance for training those models in \Cref{sec:training}. Finally, we conclude with a brief discussion about present challenges and promising directions for future research as well as possible obstacles that might be encountered during our streamlined decision process.

We could imagine different ways to go through this overview: An experienced RL practitioner may take a direct look at the tables, combining algorithmic properties and action-distribution families according to the needs for a concrete problem at hand or to discover new approaches and unresolved challenges. A less experienced reader might also be interested in the short, rather intuitive, explanations of properties. Even reading crisscross focusing only on some tables/sections should be possible since they are often independent of each other. Regarding the decisive steps we propose, it could even make sense to change their order for concrete tasks at hand. In particular, if some environmental properties are easier/clearer to assess, it probably makes sense to first make the ``evident'' decisions and to conduct individual experiments to complete less clear choices (e.g.\ it might be easier to assess if the environment has discrete or continuous states than to evaluate whether sample efficiency or training stability are more important).

\section{RL algorithms} \label{sec:algorithms}

\Cref{tab:algorithms} contains a list of RL algorithms and their respective properties. Algorithms with certain properties can be inappropriate in certain situations. \Cref{tab:model_based_vs_model_free,tab:hierarchical,tab:imitation,tab:distributed,tab:distributional,tab:on_policy_vs_off_policy,tab:stochastic_vs_deterministic,tab:value_based_vs_policy_based,tab:policy_based_variants,tab:value_based_variants,tab:value_function_learning,tab:entropy} explain these relationships between situations and algorithm properties and thus help choose an algorithm (based on the properties listed in \Cref{tab:algorithms}) that is appropriate for a given situation. More specifically, \Cref{tab:model_based_vs_model_free,tab:hierarchical,tab:imitation,tab:distributed,tab:distributional,tab:on_policy_vs_off_policy,tab:stochastic_vs_deterministic,tab:value_based_vs_policy_based,tab:policy_based_variants,tab:value_based_variants,tab:value_function_learning,tab:entropy} describe properties of \emph{situations} (``if''), desirable \emph{method properties} that are appropriate for respective situations (``then''), and corresponding \emph{reasons} (``because''). Having chosen a certain RL algorithm, there exist usually still various options for the action-distribution family that can often be used interchangeably under constraints imposed by the RL algorithm and its properties. We consider the possible choices under the respective constraints in \Cref{tab:value_based_policies,tab:stochastic_parametric_policies,tab:expressiveness}.


{\footnotesize

\begin{longtable}{|p{0.11\textwidth}|p{0.05\textwidth}|p{0.16\textwidth}|p{0.1\textwidth}|p{0.11\textwidth}|p{0.06\textwidth}|p{0.06\textwidth}|p{0.05\textwidth}|p{0.06\textwidth}|}

\caption{
Model-free RL algorithms and their properties. If \Cref{tab:model_based_vs_model_free} recommends a model-based algorithm instead, see corresponding literature, e.g.~\cite{moerland2021, moerland2020, wang2019, hamrick2021, plaat2020}. The process of choosing an algorithm based on the properties listed here is described step by step in Tables~\ref{tab:hierarchical}--\ref{tab:entropy}. If multiple algorithms are possible according to the selection process, we recommend looking into the papers where these algorithms are introduced as well as into comparative benchmark studies, e.g.~\cite{duan2016}. Note that this table is not exhaustive and many methods could be modified meaningfully such that they have different properties; yet, we try to provide a list of (popular) algorithms from the literature.
}
\label{tab:algorithms}\\

\hline

\tablehead{Name of algorithm} & 
\tablehead{On-/off-policy (see \Cref{tab:on_policy_vs_off_policy})} &
\tablehead{Value-based or policy-based or actor-critic (see Tables \ref{tab:value_based_vs_policy_based}, \ref{tab:value_based_variants}, \ref{tab:policy_based_variants})} &
\tablehead{Value-function learning (see \Cref{tab:value_function_learning})} &
\tablehead{Entropy regularization (see \Cref{tab:entropy})} &
\tablehead{Dis\-tri\-bu\-tion\-al (see \Cref{tab:distributional})} &
\tablehead{Dis\-tribu\-ted (see \Cref{tab:distributed})} &
\tablehead{Hier\-ar\-chi\-cal (see \Cref{tab:hierarchical})} &
\tablehead{Im\-i\-ta\-tion learning (see \Cref{tab:imitation})}\\ \hline

\endfirsthead 

\hline

\tablehead{Name of algorithm} & 
\tablehead{On-/off-policy (see \Cref{tab:on_policy_vs_off_policy})} &
\tablehead{Value-based or policy-based or actor-critic (see Tables \ref{tab:value_based_vs_policy_based}, \ref{tab:value_based_variants}, \ref{tab:policy_based_variants})} &
\tablehead{Value-function learning (see \Cref{tab:value_function_learning})} &
\tablehead{Entropy regularization (see \Cref{tab:entropy})} &
\tablehead{Dis\-tri\-bu\-tion\-al (see \Cref{tab:distributional})} &
\tablehead{Dis\-tribu\-ted (see \Cref{tab:distributed})} &
\tablehead{Hier\-ar\-chi\-cal (see \Cref{tab:hierarchical})} &
\tablehead{Im\-i\-ta\-tion learning (see \Cref{tab:imitation})}\\ \hline

\endhead 

Q-learning \cite{watkins1992} & Off & Tabular value-based with exact maximization & TD/Q($\lambda$) & No & No & No & No & No \\ \hline

\multicolumn{9}{|c|}{\centering $\downarrow$} \\ \hline

Sarsa \cite{rummery1994} & On & Tabular value-based with exact maximization & TD/Sarsa($\lambda$) & No & No & No & No & No \\ \hline

DQN \cite{mnih2015} & Off & Non-tabular value-based with exact maximization & TD & No & No & No & No & No \\ \hline

Double DQN (DDQN) \cite{vanhasselt2016} & Off & Non-tabular value-based with exact maximization & TD & No & No & No & No & No \\ \hline

Dueling DQN \cite{wang2016} & Off & Non-tabular value-based with exact maximization & TD & No & No & No & No & No \\ \hline

DRQN \cite{hausknecht2015} & Off & Non-tabular value-based with exact maximization & TD & No & No & No & No & No \\ \hline

Prioritized DQN \cite{schaul2016} & Off & Non-tabular value-based with exact maximization & TD & No & No & No & No & No \\ \hline

DQfD \cite{hester2018} & Off & Non-tabular value-based with exact maximization & TD/TD(n) & No & No & No & No & Yes \\ \hline

h-DQN \cite{kulkarni2016HierarchicalDR} & Off & Non-tabular value-based with exact maximization & TD & No & No & No & Yes & No \\ \hline

Distributional DQN (also called c51) \cite{bellemare2017} & Off & Non-tabular value-based with exact maximization & TD & No & Yes & No & No & No \\ \hline

QR-DQN \cite{dabney2018} & Off & Non-tabular value-based with exact maximization & TD & No & Yes & No & No & No \\ \hline

Rainbow \cite{hessel2017} & Off & Non-tabular value-based with exact maximization & TD(n) & No & Yes & No & No & No \\ \hline

Sequential DQN \cite{metz2017} & Off & Non-tabular value-based with approximate maximization and fixed search procedure & TD & No & No & No & No & No \\ \hline

AQL \cite{vandewiele2020} & Off & Non-tabular value-based with approximate maximization and learned search procedure & Q($\lambda$) & Per-state entropy regularization & No & Yes & No & No \\ \hline

GAN Q-learning \cite{doan2018} & Off & Non-tabular value-based with exact maximization & TD & No & Yes & No & No & No \\ \hline

\multicolumn{9}{|c|}{\centering $\downarrow$} \\ \hline

Bootstrapped DQN \cite{osband2016} & Off & Non-tabular value-based with exact maximization & TD & No & Yes & No & No & No \\ \hline

QT-Opt \cite{kalashnikov2018} & Off & Non-tabular value-based with approximate maximization and fixed search procedure & TD & No & Yes & Yes & No & No \\ \hline

R2D2 \cite{kapturowski2019} & Off & Non-tabular value-based with exact maximization & TD(n) & No & No & Yes & No & No \\ \hline

Agent57 \cite{badia2020} & Off & Non-tabular value-based with exact maximization & Retrace($\lambda$) & No & No & Yes & No & No \\ \hline

Ape-X DQN \cite{horgan2018} & Off & Non-tabular value-based with exact maximization & TD(n) & No & No & Yes & No & No \\ \hline

REINFORCE \cite{williams1992} & On & Policy-based & MC & Per-state entropy regularization & No & No & No & No \\ \hline

PCL \cite{nachum2017} & Off & Actor-critic & TD(n) & Soft Q-learning & No & No & No & Yes \\ \hline

Off-PAC \cite{degris2012} & Off & Actor-critic & GTD($\lambda$) & No & No & No & No & No \\ \hline

Reactor \cite{gruslys2018} & Off & Actor-critic & Retrace($\lambda$) & Per-state entropy regularization & Yes & Yes & No & No \\ \hline

A2C-SVPG \cite{liu2017} & On & Actor-critic & GAE($\lambda$) & Soft Q-learning & No & Yes & No & No \\ \hline

MPO \cite{abdolmaleki2018} & Off & Actor-critic & Retrace($\lambda$) & Kullback--Leibler divergence regularization & No & Yes & No & No \\ \hline

DSPG \cite{shi2019} & Off & Actor-critic & TD & Soft Q-learning & No & No & No & No \\ \hline

NFP/NBP \cite{tang2018} & Off & Actor-critic & TD & Per-state entropy regularization & No & No & No & No \\ \hline

AWR \cite{peng2019} & Off & Actor-critic & TD($\lambda$) & Kullback--Leibler divergence regularization & No & No & No & No \\ \hline

DPG \cite{silver2014} & Off & Actor-critic & TD & No & No & No & No & No \\ \hline

DDPG \cite{lillicrap2016} & Off & Actor-critic & TD & No & No & No & No & No \\ \hline

D3PG \cite{barth-maron2018} & Off & Actor-critic & TD(n) & No & No & Yes & No & No \\ \hline

D4PG \cite{barth-maron2018} & Off & Actor-critic & TD(n) & No & Yes & Yes & No & No \\ \hline

\multicolumn{9}{|c|}{\centering $\downarrow$} \\ \hline

TRPO \cite{schulman2015a} & On & Actor-critic & TD(n) & Kullback--Leibler divergence regularization & No & No & No & No \\ \hline
 
PPO \cite{schulman2017} & On & Actor-critic & GAE($\lambda$) & Per-state entropy and Kullback--Leibler divergence regularization & No & Yes & No & No \\ \hline

HiPPO \cite{li2020} & On & Actor-critic & GAE($\lambda$) & Kullback--Leibler divergence regularization & No & Yes & Yes & No \\ \hline

HiRO \cite{nachum2018} & Off & Actor-critic & TD & No & No & No & Yes & No \\ \hline

SNN4HRL \cite{florensa2017} & On & Actor-critic & TD(n) & Mutual-information regularization & No & No & Yes & No \\ \hline

A2C/A3C \cite{mnih2016} & On & Actor-critic & TD(n) & Per-state entropy regularization & No & Yes & No & No \\ \hline

ACER \cite{wang2017} & Off & Actor-critic & Retrace($\lambda$) & Per-state entropy and Kullback--Leibler divergence regularization & No & Yes & No & No \\ \hline

NAC \cite{peters2005} & On & Actor-critic & LSTD-Q($\lambda$) & No & No & No & No & No \\ \hline

ACKTR \cite{wu2017} & On & Actor-critic & TD(n) & Per-state entropy and Kullback--Leibler divergence regularization & No & No & No & No \\ \hline

TD3 \cite{fujimoto2018} & Off & Actor-critic & TD & No & No & No & No & No \\ \hline

SAC \cite{haarnoja2018} & Off & Actor-critic & TD & Soft Q-learning & No & No & No & No \\ \hline

Normalized actor-critic (sometimes also abbreviated as NAC) \cite{gao2018} & Off & Actor-critic & TD & Soft Q-learning & No & No & No & Yes \\ \hline

Hybrid SAC \cite{delalleau2019} & Off & Actor-critic & TD & Soft Q-learning & No & No & No & No \\ \hline

Distributional SAC \cite{duan2021} & Off & Actor-critic & TD & Soft Q-learning & Yes & Yes & No & No \\ \hline

\multicolumn{9}{|c|}{\centering $\downarrow$} \\ \hline

GAE-TRPO \cite{schulman2015b} & On & Actor-critic & GAE($\lambda$) & Kullback--Leibler divergence regularization & No & No & No & No \\ \hline

IMPALA \cite{espeholt2018} & Off & Actor-critic & V-trace(n) & Per-state entropy regularization & No & Yes & No & No \\ \hline

SEED \cite{espeholt2020} & Off & Actor-critic & V-trace(n) & Per-state entropy regularization & No & Yes & No & No \\ \hline

Ape-X DPG \cite{horgan2018} & Off & Actor-critic & TD(n) & No & No & Yes & No & No \\ \hline

Q-Prop \cite{gu2017} & Off & Actor-critic & GAE($\lambda$) & Kullback--Leibler divergence regularization & No & No & No & No \\ \hline

(RE)LAX \cite{grathwohl2018} & On & Actor-critic & MC/TD & Per-state entropy regularization & No & No & No & No \\ \hline

REDQ \cite{chen::redq} & Off & Actor-critic & MC/TD & No & No & No & No & No \\ \hline

TD-MPC \cite{hansen2022temporal} & On & Actor-critic & TD & No & No & No & No & No \\ \hline

\end{longtable}
}

\subsection{Model-free vs.\ model-based reinforcement learning} \label{sec:model_based_vs_model_free}

Considerations for deciding between model-free and model-based RL are given in \Cref{tab:model_based_vs_model_free} and explanations of terms are provided in the following.

RL algorithms differ in how they deal with the \emph{environment dynamics}, i.e.\ with the rewards $r_t$ and state-transition probabilities $p(s'\mid s,a)$. The options to choose from are:
\begin{itemize}
    \item \textbf{Model-free} RL algorithms aim to learn a policy and/or a value function directly from interacting with the environment without being aware of the underlying environment dynamics. That is, a model of the environment dynamics is not involved at any point in the algorithmic procedure. 
    \item \textbf{Model-based} RL algorithms model the environment dynamics explicitly, i.e.\ they include an \emph{environment model} somehow in the algorithmic procedure. This model can either be given in advance, or it can be learned simultaneously with the policy. The variety of model-based approaches ranges from methods that slightly augment an otherwise model-free RL approach with an environment model (e.g.\ simulating environmental transitions from the environment model in addition to real-world experience) to methods that infer an entire policy from \emph{planning} based on the environment model. 
\end{itemize}

In the following, we will primarily deal with the selection of model-free RL algorithms. In situations for which \Cref{tab:model_based_vs_model_free} recommends using model-based methods, an overview of model-based RL methods and ways to incorporate a model into RL algorithms can be found in recent surveys and overview papers \cite{moerland2021, moerland2020, wang2019, hamrick2021, plaat2020}. For an introduction to pure planning algorithms, in the sense of searching for the best action sequence in the space of all possible action sequences based on a given environment model (i.e.\ model-based but not RL due to lack of trainable parts), see for example~\cite{lavalle2006}.

\begin{table} 
    \centering
    \footnotesize
    \caption{
    Considerations for the decision between model-based and model-free reinforcement learning algorithms. The relevant method properties are described in \Cref{sec:model_based_vs_model_free}. In simulated environments, the environment dynamics (i.e.\ the rewards~$r_t$ and state-transition probabilities~$p(s'|s,a)$) are often \emph{given}. In real-world environments, the environment dynamics are often not known precisely. \emph{Training stability} refers to a training procedure in which the agent's performance improves (mostly) monotonically and which is (largely) independent of random seeds. \emph{Asymptotic performance} is the agent's performance at test time after training losses/metrics have converged. \emph{Data efficiency} refers to the efficiency of the training process with respect to the experience collected by the agent (i.e.\ agent performs well despite having little experience). In the context of RL, \emph{transfer learning} usually means exploiting knowledge about one environment to speed up the learning process in another environment (for example with the same dynamics but different reward function~\cite{moerland2021}). \emph{Safety} concerns usually deal with potentially unwanted/dangerous side-effects of an agent's behavior. \emph{Explainability} means that reasons for actions executed by the RL agent are accessible.
    }
    \label{tab:model_based_vs_model_free}
    \begin{tabularx}{\textwidth}{|b{0.2\textwidth}|b{0.2\textwidth}|b{0.1\textwidth}|b{0.393\textwidth}|}
        \hline
        \multicolumn{2}{|b{0.4\textwidth}|}{\tablehead{``If ...'' (situation)}} & & \\ 
        \cline{1-2}
         \tablehead{Environment dynamics consisting of rewards $r_t$ and state-transition probabilities $p(s'|s,a)$} & 
         \tablehead{Primary goal/concern of the learning process} &
        \tablehead{``then ...'' (method property)} & 
        \tablehead{``because ...'' (reason for method property)}
        \\ \hline
    \end{tabularx}
    \begin{tabularx}{\textwidth}{|p{0.2\textwidth}|p{0.2\textwidth}|p{0.1\textwidth}|p{0.393\textwidth}|}
        
        Not given, but feasible to learn 
        &
        Training stability and/or asymptotic performance &
        Two options (priority according to the given order):
        \begin{minipage}[t]{\linewidth}
            \begin{titemize}
                 \item Model-free
                 \item Model-based
            \end{titemize}
        \end{minipage} &
            \begin{titemize}
                 \item There exist algorithms that incorporate a learned model while maintaining good training stability and delivering excellent results, e.g.\ MuZero \cite{schrittwieser2020}. This indicates that model-based approaches can work well for the described situation.
                 \item Learning an environment model simultaneously with a policy may increase the risk of training instability and could lower asymptotic performance \cite{moerland2021}, which probably gives model-free approaches a higher priority in the described situation. 
            \end{titemize}
        \\ \hline
        
        Not given, but feasible to learn & 
        Data efficiency, transfer learning, safety, and/or explainability &
        Model-based &
        \begin{titemize}
        \item A learned model in a model-based approach helps to increase data efficiency in the training process and it is also known to be beneficial in the context of transfer learning, safety guarantees, and explainability of the algorithm \cite{moerland2021, plaat2020, hamrick2021}. 
        \end{titemize}
         \\ \hline
         
        Not given and infeasible to learn (e.g.\ environment dynamics governed by chaos) &
        Any &
        Model-free &
         \begin{titemize}
        \item In an environment where learning a model is infeasible, only a model-free approach is applicable. More precisely, it is to be expected that any model representing such an environment is deficient and, therefore, likely to be misleading. Yet, a good policy might not require knowledge about all of the environmental dynamics 
        and can, hence, potentially be learned with a model-free RL algorithm.
        \end{titemize}
        \\ \hline
        
        Given &
        Any &
        Model-based 
        &
        \begin{minipage}[t]{\linewidth}
            \begin{titemize}
                 \item The information of the environment model is usually beneficial in many regards (e.g.\ data efficiency, stability) without significant disadvantages if this model is provided in advance \cite{moerland2021}. 
                 \item If using the given model is computationally expensive or learning a sufficiently good policy does not require knowledge about environmental details, it might be beneficial to work with a simplified version of the given model. 
            \end{titemize}
        \end{minipage}
                 
    \\ \hline
        
    \end{tabularx}
  
\end{table}

\subsection{Hierarchical RL} \label{sec:hierarchical}

Selection criteria for and against a hierarchical approach can be found in \Cref{tab:hierarchical} and a short introduction to the terms is provided in the following. 

\begin{itemize}
    \item In hierarchical RL, the agent performs a coarse-to-fine cascade of several consequent decisions about the next action. The hierarchy of these decisions is often based on temporal and/or spatial abstractions that are hardwired by the programmer or learned by the agent. It is probably most convenient to consider a concrete example: Assume we have a robot that can move around with multiple wheels. Then, a high-level action (sometimes also called \emph{skill} in this context) could be ``driving an S-curve'' while a lower-level action describes how exactly to do that, i.e.\ how to turn which wheel. Consequently, multiple different policies     with different levels of abstraction can be trained. In some cases, it is even possible to transfer skills from one task to another while training other skills from scratch or to fine-tune them. In the above example, a hierarchical RL approach would allow to first train the skill ``driving an S-curve'' independently of other skills and then, potentially in a new environment, train a higher-level policy that chooses among multiple skills, e.g.\ ``driving an S-curve'' and ``driving straight ahead''.
    
    \item \textbf{Non-hierarchical} algorithms do not contain such a hierarchical structure explicitly (although parts of the neural network might learn to approximately perform a hierarchy of decisions). In the literature about hierarchical RL, non-hierarchical RL algorithms are sometimes called \emph{algorithms with flat policy}~\cite{li2020}.
\end{itemize}

\begin{table} 
    \centering
    \footnotesize
    \caption{
    Considerations for the application of a hierarchical RL method. The relevant method properties are explained in \Cref{sec:hierarchical}. If a hierarchical approach is appropriate, multiple action-distribution families (see \Cref{sec:policies}) can be selected at different levels of abstraction and/or for different skills.}
    \label{tab:hierarchical}
    \begin{tabularx}{\textwidth}{|b{0.3\textwidth}|b{0.2\textwidth}|b{0.42\textwidth}|}
    \hline
     
    \multicolumn{1}{|b{0.3\textwidth}|}{\tablehead{``If ...'' (situation)}} & &\\ 
    \cline{1-1}

    \tablehead{Complexity of action sequences} &
    \tablehead{``then ...'' (method property)} & 
    \tablehead{``because ...'' (reason for method property)}  \\ \hline

    \end{tabularx}

    \begin{tabularx}{\textwidth}{|p{0.3\textwidth}|p{0.2\textwidth}|p{0.42\textwidth}|}

    Highly complex action sequences or action sequences that could be divided hierarchically into sub-routines at different levels of granularity & 
     Hierarchical approach (see for example Ref.~\cite{pateria2021} for good practices for design of hierarchical RL) & 
        \begin{titemize}
            \item Highly complex action sequences can often be described more easily on different hierarchical levels, e.g., for a robot, a high-level action could be ``grab the object,'' and a low-level sequence of actions could be ``move fingers appropriately''.
            \item Many methods not primarily designed for hierarchical actions perform poorly when applied to hierarchical tasks \cite{duan2016}.
        \end{titemize}
        \\ \hline

    Action sequences are at most moderately complex & 
    Non-hierarchical approach &
        \begin{titemize}
            \item A hierarchical approach is probably not very helpful if action sequences are not too complex.  
            \item Hierarchical approaches usually require more advanced training procedures and architectures than common RL algorithms.
        \end{titemize}
    \\ \hline
    
    \end{tabularx}
\end{table}

\subsection{Imitation learning} \label{sec:imitation_learning}

\Cref{tab:imitation} summarizes the considerations about when to apply imitation learning in an RL task. Terms related to imitation learning are explained in the following.

\begin{itemize}
\item A field closely related to reinforcement learning is \textbf{imitation learning}. While RL algorithms aim to solve a decision-making problem based on some environmental feedback (the rewards~$r$), imitation-learning algorithms learn a policy based on observing the behavior of an expert. While a detailed overview of imitation-learning algorithms is beyond the scope of this work (see related literature about imitation learning, e.g.~\cite{argall2009, torabi2019, zhu2020, ho2016, kostrikov2020, ross2011}), our \Cref{tab:algorithms} solely lists methods that combine imitation learning with RL, as well as RL methods that do not perform imitation learning at all, but our table does not list methods that perform imitation learning without RL. Notably, the performance of an agent trained with pure imitation learning is quite bounded by the performance of the expert. In contrast, an algorithm that uses both concepts (RL \emph{and} imitation learning) jointly might have better chances of outperforming the expert at the end of the training phase \cite{hester2018}.

Joint reinforcement and imitation learning usually means to pre-train an agent first solely on an imitation loss based on the (dis)similarity between the behavior of an expert and the RL agent, and subsequent reward-based training (as in common RL) with an additional imitation loss term. Theoretically, it would be possible to remove the imitation loss after the imitation pre-training phase, but as the authors of \cite{hester2018} argue, this seems to be sub-optimal.

\item Algorithms \textbf{without imitation learning} listed in \Cref{tab:algorithms} usually do not use expert trajectories in the learning process of the agent.

\end{itemize}

Note that, even though it might be tempting to simply augment a replay buffer in an off-policy algorithm (cf.~\Cref{sec:definition-on-off-policy}) with expert trajectories, this can have a devastating impact on the training process without further modifications \cite{fujimoto2019} but can also work well \cite{nachum2017}. If such augmentation is an option, e.g.\ because the expert trajectories do already exist and an off-policy algorithm with replay buffer seems to be a good choice as per \Cref{tab:on_policy_vs_off_policy}, a closer look into the literature about the respective algorithm and/or additional experiments might be required.

\begin{table} 
    \centering
    \footnotesize
    \caption{Considerations for and against including imitation learning in the RL algorithm. The relevant method properties are described in \Cref{sec:imitation_learning}. An \emph{expert} can be, for example, a human or a technical system that is known to solve the task well.}
    \label{tab:imitation}
    \begin{tabularx}{\textwidth}{|b{0.2\textwidth}|b{0.2\textwidth}|b{0.52\textwidth}|}
    \hline
    
    \multicolumn{1}{|b{0.2\textwidth}|}{\tablehead{``If ...'' (situation)}} & & \\
    \cline{1-1}
    \tablehead{Availability of an expert} &
    \tablehead{``then ...'' (method property)} & 
    \tablehead{``because ...'' (reason for method property)}  \\ 
        \hline
    \end{tabularx}
    \begin{tabularx}{\textwidth}{|p{0.2\textwidth}|p{0.2\textwidth}|p{0.52\textwidth}|}
        No expert available &
        No imitation learning &
        If no expert is available, imitation learning is not possible.
        \\ 
        \hline
        
        An expert is available &
        Imitation learning
        &
            \begin{titemize}
                \item Learning from the behavior of an expert is likely to be beneficial for the training process itself (e.g.\ better stability, faster convergence) and may improve test performance \cite{hester2018}. The overhead in terms of computation and implementation effort is typically low if expert trajectories are already available.
                \item Even though the imitation loss could be removed after some initial pre-training phase, this seems to be sub-optimal \cite{hester2018}. 
            \end{titemize}
        \\
        \hline
       
    \end{tabularx}
\end{table}

\subsection{Distributed algorithms}\label{sec:distributed}
Considerations for deciding for or against a \textbf{distributed} (not to be confused with distributional) RL algorithm are given in \Cref{tab:distributed} and an explanation of this term is provided in the following. The two options to choose from are:
\begin{itemize}
    \item In \textbf{distributed} algorithms, there exist \emph{two or more actors}, i.e.\ parallel sub-processes of which each acts in a separate instance of the environment and creates training data, and \emph{one or more} learners, i.e.\ parallel sub-processes that are responsible for updating parameters, at training time. The number of learners can be smaller, equal, or larger than the number of actors. Especially if the acquisition of experience is slow, the parallel execution of actor processes can significantly reduce training time since more experience is acquired in less time. Multiple learners allow, for example, for an asynchronous and parallel computation of gradients \cite{mnih2016}, thus also speeding up training. 
    \item \textbf{Non-distributed} algorithms only employ one actor and one learner (see the definitions in the previous bullet point).
\end{itemize}

\begin{table} 
    \centering
    \footnotesize
    \caption{Considerations for and against a distributed algorithm. The relevant method properties are described in \Cref{sec:distributed}. 
    }
    \label{tab:distributed}
    \begin{tabularx}{\textwidth}{|b{0.3\textwidth}|b{0.2\textwidth}|b{0.42\textwidth}|}
    \hline
    \multicolumn{1}{|b{0.3\textwidth}|}{\tablehead{``If ...'' (situation)}} & & \\
    \cline{1-1}
    \tablehead{Computational resources (consisting of computation units and memory)} &
    \tablehead{``then ...'' (method property)} & 
    \tablehead{``because ...'' (reason for method property)}  \\ 
        \hline
    \end{tabularx}
    \begin{tabularx}{\textwidth}{|p{0.3\textwidth}|p{0.2\textwidth}|p{0.42\textwidth}|}
        
        There is no way of running two or more actors in parallel during training (e.g., only one computation unit is available). & 
        Not distributed &
        \begin{titemize}
            \item Learning in a distributed manner requires at least enough computational resources to run two actors in parallel. 
        \end{titemize}
        
        \\ 
        \hline
        
        It is possible to run several actors (and maybe also learners) in parallel during training (e.g.\ several computation units available). 
        &
        Two options (priority according to the given order):
            \begin{titemize}
                \item Distributed, optionally learning a diverse set of policies
                \item Not distributed
            \end{titemize}
            &
            \begin{titemize}
                \item Learning in a distributed manner by acting in multiple environment instances in parallel and exchanging data usually reduces training time. (We assume a training procedure with fixed hyperparameters. On the other hand, during hyperparameter tuning, it might be more efficient to try different hyperparameters on different computation units instead of splitting up each experiment.)
                 \item Non-distributed algorithms can still be used but are potentially less preferred as running several actors is impossible.
            \end{titemize}
            \\
        \hline
       
    \end{tabularx}
\end{table}

\subsection{Distributional algorithms} \label{sec:distributional}

Considerations to decide for or against a \textbf{distributional} (not to be confused with distributed) algorithm are in \Cref{tab:distributional} and related terms are explained in the following.

\begin{itemize}
    \item \textbf{Distributional approaches} estimate the probability distribution of the cumulative reward $\sum_t r_{t}$ instead of merely its expected value $\E[\sum_t r_{t}]$ (i.e.\ the value function), where $r_{t}$ is the reward at time $t$. Distributional approaches are usually combined with value-based or actor-critic methods since pure policy-based methods do not learn a value function (and value-based and actor-critic methods do). 
    \item \textbf{Non-distributional} approaches only consider the expectation of the cumulative reward as introduced previously in \Cref{sec:value_function_learning}. 
\end{itemize}

\begin{table} 
    \centering
    \footnotesize
    \caption{Considerations for the selection of a distributional algorithm. The relevant method properties are described in \Cref{sec:distributional}. The \emph{risk} of a certain action could be understood as the probability of observing a low cumulative reward (in the context of the environment) when taking that action at a certain state and following the current policy $\pi$ afterwards. 
    The actually observed cumulative reward may in general be subject to randomness introduced by a stochastic policy or a stochastic environment.}
    \label{tab:distributional}
    \begin{tabularx}{\textwidth}{|b{0.15\textwidth}|b{0.2\textwidth}|b{0.57\textwidth}|}
    \hline
    \multicolumn{1}{|b{0.15\textwidth}|}{\tablehead{``If ...'' (situation)}} & &\\
    \cline{1-1}
    \tablehead{Estimating the ``risk'' of taking a certain action} & 
    \tablehead{``then ...'' (method property)} & 
    \tablehead{``because ...'' (reason for method property)}  \\ \hline

    \end{tabularx}
    \begin{tabularx}{\textwidth}{|p{0.15\textwidth}|p{0.2\textwidth}|p{0.57\textwidth}|}

    Required & 
    Distributional & 
        \begin{titemize}
            \item Only distributional model-free methods allow for risk estimation since they learn a distribution over the cumulative reward $\sum_t r_{t}$ instead of the expected value $\E[\sum_t r_t]$, where $r_t$ is short for the reward at time $t$.
            \item Distributional approaches may also improve exploration generalization, training stability, and overall performance at the cost of training an additional density estimator \cite{henderson2017,osband2016}.
        \end{titemize}
        \\ \hline
    
    Not required & 
    Low priority (in the sense of ``the decision has lower priority than other decisions ''): distributional 
    & 
        \begin{titemize}
            \item If risk estimation is not required, a distributional approach is not necessary.
            \item A distributional method can still be beneficial as described above.
        \end{titemize}
        \\
    \hline
    
    \end{tabularx}
\end{table}

\subsection{On-policy vs.\ off-policy learning} \label{sec:on_policy_vs_off_policy}
Considerations for deciding between on- and off-policy algorithms are given in \Cref{tab:on_policy_vs_off_policy} and explanations of these terms are provided in the following.

\subsubsection{Target policy, behavior policy, test-time policy} \label{sec:policy_definitions}

We define the terms \emph{target policy}, \emph{behavior policy}, and \emph{test-time policy} as follows:
\begin{itemize}
    \item The \textbf{target policy} (at each given timepoint during training) is a policy that is being trained, and from which the behavior policy and the test-time policy are computed as explained below. The term \emph{target policy} is not to be confused with the term \emph{output target} (values that a neural network tries to learn to output, but does not output precisely).
    
    \item The \textbf{behavior policy} (at each given timepoint during training) is the policy that is used to perform actions during the training of the target policy. While the target policy is an attempt to maximize the cumulative rewards given the current, typically incomplete, knowledge about the environment, the behavior policy can differ from it in terms of additional noise (in order to increase exploration, thus gathering more knowledge about the environment and possibly discovering better behaviors) and/or in terms of rarer updates (in order to increase training stability: the expected cumulative reward depends on future actions and, hence, can be difficult to estimate if the behavior policy is changed too frequently and there is no consistency in the actions). The behavior policy is from time to time (for example continuously or in fixed time intervals) ``computed'' from the target policy. Examples of such occasional updates of the behavior policy are:
    \begin{itemize}
        \item The behavior policy can be, from time to time, set equal to the current target policy.
        \item The behavior policy can be, from time to time, set equal to the current target policy \emph{with additional noise}. The purpose of the additional noise is to improve exploration.
        \item The behavior policy can be, from time to time, set equal to the current target policy \emph{with the addition of another trained part}, for example, noise with trainable parameters (such as mean and variance).
    \end{itemize}
    
    \item The \textbf{test-time policy} is the policy used at test time, i.e.\ the ``final recommendation''. While the final target policy (the target policy at the end of training) is an attempt to perform optimally given the knowledge gathered about the environment, that training attempt might be imperfect, and/or the knowledge might be incomplete. In order to, for example, circumvent new kinds of obstacles in the environment that were not encountered during training, the test-time policy might have additional noise compared to the final target policy. The test-time policy can be computed from the final target policy for example in one of the following ways:
    \begin{itemize}
        \item The test-time policy can be set equal to the final target policy.
        \item The test-time policy can be set equal to the final target policy \emph{with additional noise}.
        \item If the target policy is stochastic (cf.~\Cref{tab:stochastic_vs_deterministic}), a deterministic test-time policy can be computed from it, e.g.\ by choosing the mode of the respective probability distribution~\cite{haarnoja2018}. This is most suitable if exploration should be removed entirely at test time. 
    \end{itemize}
    Note that any modification to the target policy at test-time may degrade the agent's performance and, hence, should only be considered if it cannot be learned/observed by the agent. Moreover, since such post hoc modifications are usually not part of model-free RL algorithms, we assume in this work that the test-time policy and the final target policy are equal unless stated otherwise.
\end{itemize}

\subsubsection{Stochastic vs.\ deterministic target policy}\label{sec:stochastic_vs_deterministic}

Considerations for deciding between a stochastic and a deterministic target policy are given in \Cref{tab:stochastic_vs_deterministic} and explanations of these terms are provided in the following.

\begin{itemize}
    \item A \textbf{stochastic policy} is defined by a conditional probability distribution $\pi_{\theta}(a\mid s)$, where $\theta$ are the learnable parameters. Given a state~$s$, the action~$a$ is sampled from this probability distribution. The probability distribution over actions (conditional, given the state) can differ from state to state.
    
    \item A \textbf{deterministic policy} is defined by a deterministic function $\mu_\theta(s)$ that maps each state $s$ to an action, where $\theta$ are the learnable parameters. Note that a deterministic policy is a special case of a stochastic policy where $\pi_\theta(a\mid s) = 1$ if $a = \mu_\theta(s)$.
\end{itemize}

\begin{table}
    \centering
    \footnotesize
    \caption{
    Considerations for the decision between a stochastic and a deterministic target policy. The relevant definitions and properties are introduced in \Cref{sec:stochastic_vs_deterministic}. Rewards are \emph{sparse} if they are only obtained in a few states. Rewards are \emph{dense} if a reward is obtained in every state. \deceptive{} Proofs that an optimal policy is deterministic in certain situations can be found in~\cite{puterman2005}; for example, in fully observable stationary environments (i.e.\ environments where environment dynamics and reward functions do not change over time) with a finite state space, finite action space, and bounded reward function, there always exists a deterministic optimal policy~\cite[Theorem~6.2.10]{puterman2005}. On the other hand, the optimal policy could be stochastic if states are not fully observable or a deterministic behavior is generally disadvantageous, e.g.\ in a game where competitive players could exploit deterministic behavior. 
    }
    
    \label{tab:stochastic_vs_deterministic}
    \begin{tabularx}{\textwidth}{|b{0.1\textwidth}|b{0.08\textwidth}|b{0.1\textwidth}|b{0.15\textwidth}|b{0.437\textwidth}|}
        \hline
  
        \multicolumn{3}{|b{0.28\textwidth}|}{\tablehead{``If ...'' (situation)}} & &\\
        \cline{1-3}
        \multicolumn{2}{|b{0.18\textwidth}|}{\tablehead{Environment properties}} & 
        \multicolumn{1}{|b{0.1\textwidth}|}{\tablehead{Method properties}} & &\\
        \cline{1-3}
        \tablehead{Rewards} &
        \tablehead{Optimal policy $\pi^{*}(a\mid s)$} &
        \tablehead{On vs.\ off-policy (see \Cref{tab:on_policy_vs_off_policy})} &
        \tablehead{``then ...'' (method property)} & 
        \tablehead{``because ...'' (reason for method property)} \\
        \hline

    \end{tabularx}
        \begin{tabularx}{\textwidth}{|p{0.1\textwidth}|p{0.08\textwidth}|p{0.1\textwidth}|p{0.15\textwidth}|p{0.437\textwidth}|}
        
        Non-deceptive and (quite) dense & 
        Probably deterministic & 
        Off-policy &
        Two options (with equal priority):
            \begin{titemize}
                \item Stochastic target policy
                \item Deterministic target policy
            \end{titemize} 
            &
            \begin{titemize}
                \item Stochastic policies can become (as a result of training) or be chosen (manually, as a test-time policy) close to deterministic if this is desirable. Therefore, there is often no need to select a deterministic target policy.
                \item Off-policy algorithms allow for decoupling the behavior from the target policy; learning a deterministic target policy while improving exploration by following a stochastic behavior policy is, in particular, interesting if the optimal policy is expected to be deterministic.
                \item There is probably no clear choice between the two listed possibilities in such environments, see for example \cite[Appendix E]{haarnoja2018}, where a deterministic policy performs similarly to a stochastic one on all tasks but one.
            \end{titemize}
        \\ \hline
        
        Non-deceptive and (quite) dense & 
        Probably deterministic & 
        On-policy &
        Stochastic target policy
        &
            \begin{titemize}
                \item According to the above row, a stochastic or a deterministic target policy could be suitable for the given environmental situation. 
                \item However, a deterministic target policy in an on-policy algorithm impedes exploration at training time, which is usually undesirable. Therefore, we conclude that a stochastic target policy is probably the best choice in the given situation. 
            \end{titemize}
        \\ \hline
        
        Sparse, deceptive, or unclear in advance & 
        Probably deterministic &
        Any &
        Stochastic target policy & 
            \begin{titemize}
                \item Deterministic policies might not work well if rewards are sparse, even if the optimal policy is deterministic \cite{matheron2019}.
                \item Randomness also increases robustness against deceptive rewards and/or local minima \cite{mazoure2020}.
            \end{titemize}
        \\ \hline
        
        Any & 
        Probably stochastic  & 
        Any &
        Stochastic target policy & 
        \begin{titemize}
            \item A deterministic policy cannot model stochastic behavior.    
        \end{titemize}
        \\ \hline

    \end{tabularx}
  
\end{table}

\subsubsection{Definition of on-policy and off-policy learning} \label{sec:definition-on-off-policy}

Having the above distinction between target and behavior policy in mind, the difference between on- and off-policy algorithms can be summarized as follows:

\begin{itemize}
    \item In \textbf{on-policy} algorithms, the behavior policy and the target policy are always equal and only experience collected with the \emph{current} target/behavior policy is used to update the target policy. That is, the experience used for updating (``learning about'') the target policy has been collected using the target policy in its current state (right before the parameter update). After the parameter update, the previously collected experience becomes worthless for the on-policy algorithm since the policy under which the experience has been acquired is then different from the current target policy, and new experience with the updated target policy needs to be collected.
    
    \item \textbf{Off-policy} algorithms allow the behavior policy to differ from the target policy. In contrast to on-policy methods, experience collected under a behavior policy that was different than the current target policy can be used for updating the parameters of the target policy. A typical example is to increase the amount of exploration by using a behavior policy that has more randomness than the current target policy. Also, the allowed discrepancy between the two policies is often exploited by using a \textbf{replay buffer}, in which experience acquired under the behavior of multiple older versions of the behavior policy is stored. It is also possible to integrate some expert trajectories into the replay buffer, leading to a form of imitation learning, see also \Cref{sec:imitation_learning}. Finally, the experience in a replay buffer can be \textbf{prioritized} such that training samples that seem to be promising for the training process are replayed more frequently than potentially less important ones \cite{schaul2016}. 
    
\end{itemize}

Note that it is no problem to use experience collected under the current policy (``on-policy data'') in off-policy algorithms, see, e.g.\ \cite{nachum2017, kalashnikov2018}. In contrast, the opposite (using off-policy data for training in on-policy algorithms) is typically not possible.

\begin{table} 
    \centering
    \footnotesize
    \caption{
    Considerations for the decision between on-policy and off-policy algorithms. The relevant method properties are described in \Cref{sec:definition-on-off-policy}. Training is considered \emph{stable} if the performance of the agent in terms of the cumulative rewards is increasing (mostly) monotonically and results are largely independent of random seeds. An \emph{unstable} training procedure does not necessarily lead to bad results but more extensive hyperparameter tuning or multiple training runs might be required compared to a stable training procedure. Experience could, for example, be \emph{cheap} if the agent acts in a fast simulation. On the other hand, real-world physical experience is often more \emph{costly} to acquire.
    }
    \label{tab:on_policy_vs_off_policy}
    \begin{tabularx}{\textwidth}{|b{0.15\textwidth}|b{0.08\textwidth}|b{0.06\textwidth}|b{0.11\textwidth}|b{0.11\textwidth}|b{0.33\textwidth}|}
    \hline
    
    \multicolumn{4}{|b{0.4\textwidth}|}{\tablehead{``If ...'' (situation)} } &  &  \\
    \cline{1-4}
     \multicolumn{3}{|b{0.29\textwidth}|}{\tablehead{Environment properties}} & 
     \tablehead{Method properties} & & \\ 
     \cline{1-4}
        \tablehead{Exploration} & 
        \tablehead{Training stability} &
        \tablehead{New experience} &
        \tablehead{Target policy stochastic vs.\ deterministic (see \Cref{tab:stochastic_vs_deterministic})} &
        \tablehead{``then ...'' (method property)} &
        \tablehead{``because ...'' (reason for method property)}
          \\ 
        \hline

        \end{tabularx}

        \begin{tabularx}{\textwidth}{|p{0.15\textwidth}|p{0.08\textwidth}|p{0.06\textwidth}|p{0.11\textwidth}|p{0.11\textwidth}|p{0.33\textwidth}|}
       
        Not too important (e.g.\ because the percentage of good actions is high) 
        & 
        Important & 
        Cheap & 
        Stochastic
        &
        On-policy &
            \begin{titemize}
                \item On-policy methods are usually more stable than off-policy methods, especially in combination with function approximation like neural networks \cite[Chapter~11]{sutton2018}.
                \item The lower sample efficiency in on-policy methods compared to off-policy methods is not a large drawback if acquiring new experience is cheap.
                \item The stochastic target policy (which is equal to the behavior policy in on-policy algorithms) ensures exploration at training time. 
            \end{titemize}
            \\
        \hline
        
        \multicolumn{3}{|p{0.29\textwidth}|}{Good exploration and/or efficient use of experience is more important than training stability (e.g.\ because acquiring new experience is expensive)}
        &
        Any
        &
        Off-policy, with additional options:
            \begin{titemize}
                \item Experience replay
                \item Pri\-or\-i\-ti\-za\-tion of experience 
            \end{titemize}
         &
            \begin{titemize}
                \item In off-policy methods, it is possible to use a behavior policy with increased randomness compared to the target policy for good exploration.
                \item In off-policy methods, it is possible to re-use past experience in a non-volatile replay buffer for higher sample efficiency. 
                \item This flexibility often comes at the cost of lower training stability compared to on-policy methods, which can hardly be avoided \cite[Chapter 11]{sutton2018}.
                \item An optional prioritization of the samples in the replay buffer allows the agent to focus on important experience, i.e.\ tuples of the form $(s, a, s', r)$, at the cost of additional computations per experience.
            \end{titemize}
            \\ 
        \hline
        
    \end{tabularx}

\end{table}

\subsection{Value-based vs.\ policy-based vs.\ actor-critic} \label{sec:value_based_vs_policy_based}

Considerations for the decision between a value-based, policy-based, or actor-critic algorithm can be found in \Cref{tab:value_based_vs_policy_based}. If this table indicates a value-based algorithm, \Cref{tab:value_based_variants} lists considerations for deciding between the different subcategories based on the size of the state and action spaces. On the other hand, if a policy-based or actor-critic approach is required, \Cref{tab:policy_based_variants} contains considerations for the decision about a learned critic. The terminology and underlying concepts are explained in the following.

The primary goal of RL algorithms for \emph{control tasks} is to learn a policy $\pi(a\mid s)$ that maximizes the cumulative reward $\sum_t r_{t}$ (while in RL algorithms for \emph{prediction tasks}, the goal is to learn to predict the outcome of a given policy in terms of rewards for an unknown environment model). In general, RL algorithms for control tasks differ in the way the agent's policy is defined and, therefore, also in the way the policy is learned. The options with respective subcategories are:

\begin{itemize}
    \item \textbf{Value-based} algorithms infer the policy directly from a learned action-value function $Q(s,a) = \mathbb{E}(\sum_{\tau=t}^{\Tmax}r_{\tau+1} \mid  s_t = s, a_t = a)$ that assigns an expected value of subsequent cumulative rewards to every possible state-action pair (a more accurate name would be ``state-action-value function'' but ``action-value function'' is widely established). See \Cref{sec:value_function_learning} for methods to learn such an action-value function. In value-based algorithms, the agent selects an action $a_t = \argmax_a Q(s_t, a)$ at each time step $t$. Due to the deterministic computation of the $\argmax$, this leads to a primarily deterministic policy (if multiple actions maximize $Q$ at a certain state, one of them may be selected randomly). However, there exist ways to make the actions stochastic (cf.\ \Cref{sec:value_based_policies}), leading to policies that deviate from $\argmax_a Q(s_t, a)$ but are still built upon this definition and therefore considered value-based. Value-based algorithms can further be grouped into the following subcategories.
    
        \begin{itemize}
        \item \textbf{Tabular value-based} methods define an action value for each possible state-action pair $(s,a)$ in a tabular way, i.e.\ a table that represents the action-value function $Q(s,a)$. For tabular action-value-based methods (in contrast to non-tabular methods, see below), very good convergence guarantees exist and exact maximization is usually possible \cite{watkins1989}, \cite[Chapters~5,6]{sutton2018}. Tabular methods are not related to deep learning and we will not consider them further; a detailed description can be found in \cite[Chapter~I]{sutton2018}. For the sake of completeness, we indicate in \Cref{tab:value_based_vs_policy_based} in which situation a tabular value-based approach might be most appropriate. 
        
        \item \textbf{Non-tabular value-based} methods approximate the action-value function with a non-tabular function $Q_\phi(s,a)$, where $\phi$ are trainable parameters. In most cases, $Q_\phi$ is represented by a deep neural network and an action at time $t$ is given by $a_t = \argmax_a Q_\phi(s_t, a)$.
        
        We further subdivide non-tabular value-based algorithms according to the following categories that are also used by \cite{vandewiele2020}: 
            \begin{itemize}
                \item In \textbf{value-based} methods \textbf{with exact maximization} over actions, the action $a$ is given by the one that exactly maximizes $Q_\phi(s,a)$ at the current state $s$.
                
                \item \textbf{Value-based} methods \textbf{with approximate maximization and fixed search procedure} compute an action that might not maximize the action-value function exactly, using a fixed procedure such as the cross-entropy method \cite{rubinstein2004}. 
                
                \item \textbf{Value-based} methods \textbf{with approximate maximization and learned search procedure} compute an action (that might not maximize the action-value function exactly) using another learned function that proposes a set of promising actions \cite{vandewiele2020}. The idea is to reduce the potentially large set of possible actions to a smaller set of which the best action can be easily determined. Such methods are conceptually similar to actor-critic methods (defined below). The difference is that the executed action is determined by the action-value function from the set of proposals in a value-based method with approximate maximization and a learned search procedure. In contrast, in actor-critic methods, an action proposed by the actor is directly executed without having its action value assessed before.  
            \end{itemize}
        
        \end{itemize}
        
    \item \textbf{Policy-based} algorithms, on the other hand, parameterize the policy explicitly. That is, they have a set of learnable parameters $\theta$ defining the policy $\pi_\theta(a\mid s)$. At each time step $t$, the agent selects an action $a_t \sim \pi_\theta(a\mid s_t)$. Note that the policy does not necessarily need to be stochastic --- it can also be a deterministic function $a = \mu_\theta(s)$, i.e.\ $\pi(a\mid s) = 1$ iff $a=\mu_\theta(s)$. We ignore this detail for the sake of simplicity unless stated otherwise.
    
    Purely policy-based methods (in contrast to actor-critic methods, see next bullet point), also called REINFORCE algorithms due to~\cite{williams1992}, assess the quality of the current policy based on cumulative rewards $\sum_t r_{t}$ obtained from episodes that have been completed with this policy. Similarly to Monte Carlo value-function learning (see \Cref{sec:value_function_learning}), this implies that parameters can only be updated at the end of completed episodes. In order to increase training stability, a learned baseline, often a state-value function $V(s)$ trained with one of the algorithms of \Cref{sec:value_function_learning}, is commonly subtracted from the observed cumulative reward in such methods \cite{williams1992,mnih2016}, see \Cref{sec:stability}. Policy-based methods (without a learned critic) allow for unbiased estimates of the policy gradients, 
    in contrast to actor-critic methods \cite{williams1992} (see next bullet point). Note that we ignore estimates of higher-order derivatives like the Hessian and only consider policy gradients for simplicity.

    \item \textbf{Actor-critic} methods can be considered to be both policy-based and value-based. These algorithms update the policy parameters based on an additional learned value function that is used in place of actually observed cumulative rewards in policy-based methods. This function is commonly called the \emph{critic} and the policy function is commonly called an \emph{actor} in the context of actor-critic methods. The critic approximates the state-value function $V(s)$ (future cumulative reward as an expectation over all actions in state $s$), the action-value function $Q(s,a)$ (future cumulative reward given a particular action $a$ in state $s$), or the advantage function $A(a,s) = Q(s, a) - V(s)$ (cumulative reward with subtracted state-dependent baseline \cite{mnih2016}, i.e.\ the advantage of taking the \emph{specific} action $a$ compared to the expectation over all possible actions, where action probabilities are given by the policy $\pi(a\mid s)$).
    
    The critic can be trained with one of the methods of \Cref{sec:value_function_learning}, similar to training to approximate the value function in value-based methods. The actor, on the other hand, can be trained such that the value of its policy, as assessed by the critic, is highest. However, the critic usually provides only an approximation of the true objective (i.e.\ of the expected future cumulative reward). As a consequence of this approximation, the gradient of the critic's outputs with respect to policy parameters (for example, the parameters of the NN that represents the policy function), which is usually referred to as the \emph{policy gradient}, is typically biased in actor-critic methods. This bias implies that a perfectly trained actor might maximize the critic's outputs but not necessarily the cumulative rewards in the actual environment. Yet, policy gradients in actor-critic methods often have lower variance compared to policy-based methods without a learned critic \cite[Section~13.8]{sutton2018}. A low variance in gradient estimates is known to be favorable for the number of training steps required to train a good agent~\cite{greensmith2004}. 
    
    Note that actions in actor-critic methods are obtained from a freely parameterized policy function without being ``assessed'' by an action-value function before execution. This assessment distinguishes actor-critic methods from value-based methods with approximate maximization and learned search procedure (see above).

\end{itemize}

\begin{table} 
    \centering
    \footnotesize
    \caption{
    Considerations for the decision between value-based, policy-based, and actor-critic algorithms. The relevant method properties are described in \Cref{sec:value_based_vs_policy_based}. The mentioned considerations for the value functions in this table refer to the state-value/action-value/advantage function $V_{\pi^{*}}$/$Q_{\pi^{*}}$/$A_{\pi^{*}}$ (summarized by 'value functions') under an optimal policy $\pi^{*}$. In general, the shape of the value functions may be different under an arbitrary policy $\pi$, but $\pi^{*}$ is usually approached during training and, therefore, most relevant.
    }
    \label{tab:value_based_vs_policy_based}
    
    \begin{tabularx}{\textwidth}{|b{0.19\textwidth}|b{0.12\textwidth}|b{0.18\textwidth}|b{0.405\textwidth}|}
    \hline
  
    \multicolumn{2}{|b{0.31\textwidth}|}{\tablehead{``If ...'' (situation)}} & &\\
    \cline{1-2}
    \tablehead{Environment properties} & 
    \tablehead{Method properties} & &\\
    \cline{1-2}
    \tablehead{Relation between true value functions and optimal policy $\pi^{*}$} &
    \tablehead{Deterministic vs.\ stochastic target policy (see \Cref{tab:stochastic_vs_deterministic})} &
    \tablehead{``then ...'' (method property)} & 
    \tablehead{``because ...'' (reason for method property)} \\
    \hline   

    \end{tabularx}

    \begin{tabularx}{\textwidth}{|p{0.19\textwidth}|p{0.12\textwidth}|p{0.18\textwidth}|p{0.405\textwidth}|}    
        
        At least one value function not expected to be considerably more complex (harder to learn) than $\pi^{*}$ (for example, the relation between an action and its value, i.e.\ the sum of subsequent rewards, is often simple in short-episode environments) &  
        Deterministic target policy & 
        All of the following are worth trying:
            \begin{titemize}
                \item Actor-critic (as per \Cref{tab:policy_based_variants})
                \item Policy-based (as per \Cref{tab:policy_based_variants})
                \item Value-based (variants as per \Cref{tab:value_based_variants})
            \end{titemize}
            &
                \begin{titemize}
                    \item  If learning (at least one) value function (i.e.\ using a value-based or an actor-critic method) and the policy function (i.e.\ using a policy-based or an actor-critic method) is similarly easy/hard, then there is no obvious reason to prioritize one approach over the other (cf.\ the last row of this table). 
                Further considerations are:
                \begin{titemize}
                    \item Value-based methods are sometimes considered more stable during training than actor-critic methods \cite{kalashnikov2018} and known to be more sample-efficient than policy-based methods \cite{nachum2017}.
                    \item Policy-based and actor-critic methods can automatically find a good trade-off between high exploration at the beginning and exploitation of good actions at the end of the training procedure, which is difficult to achieve in value-based methods \cite[Section~13.1]{sutton2018}.
                    \item Which of the above arguments weighs most cannot be generalized without further experiments. We conclude that it might be necessary to try algorithms with different properties in this situation.
                \end{titemize}
            \end{titemize}
     \\ \hline
        
        Value functions are not expected to be considerably more complex (see above) &  
        Stochastic target policy & 
        All of the following are worth trying, with potentially lower priority of the last option than the first two:
        \begin{titemize}
                \item Policy-based (as per \Cref{tab:policy_based_variants})
                \item Actor-critic (as per \Cref{tab:policy_based_variants})
                \item Value-based (variants as per \Cref{tab:value_based_variants})
            \end{titemize}
        &
        \begin{titemize}
            \item This situation is very similar to the row above. However, a desired stochastic target policy might make value-based methods lower priority compared to policy-based/actor-critic methods. The reason is that the variety of stochastic policies (e.g.\ in terms of learnable modes) realizable in value-based algorithms is more limited than in policy-based/actor-critic algorithms (cf.~\Cref{tab:value_based_policies,tab:stochastic_parametric_policies}).    
        \end{titemize}
         \\ \hline
        
        Policy  function  is expected to  be considerably simpler than (at least one of) the value functions (for example, the exact value of each state-action pair could be difficult to estimate but an optimal behavior is  rather  simple like ``always turn left in state $s_i$'')
        & 
        Any &
            \begin{titemize}
                \item Policy-based (as per \Cref{tab:policy_based_variants})
                \item Actor-critic (as per \Cref{tab:policy_based_variants})
            \end{titemize}
        
        &
        \begin{titemize}
        \item In value-based methods, an accurately learned action-value function is crucial for an optimal policy. In contrast, policy-based and actor-critic methods parameterize the policy directly, which is expected to be simpler in this situation.    
        \end{titemize}
         \\ 
        \hline
        
    \end{tabularx}
   
\end{table}

\begin{table} 
    \centering
    \footnotesize
    \caption{Considerations for the decision between a (pure) policy-based and an actor-critic RL algorithm (for a preceding decision between policy-based, actor-critic, and value-based algorithms, see \Cref{tab:value_based_vs_policy_based}). This decision between a policy-based or actor-critic method is, at its core, about using a learned critic in an otherwise policy-based RL algorithm. The relevant method properties are described in \Cref{sec:value_based_vs_policy_based}.
    }
    \label{tab:policy_based_variants}
    \begin{tabularx}{\textwidth}{|b{0.07\textwidth}|b{0.1\textwidth}|b{0.08\textwidth}|b{0.117\textwidth}|b{0.5\textwidth}|}
    \hline
    
     \multicolumn{3}{|b{0.25\textwidth}|}{\tablehead{``If ...'' (situation)}} & &\\
        \cline{1-3}
        \tablehead{Environ-ment properties} & 
        \multicolumn{2}{b{0.18\textwidth}|}{\tablehead{Method properties}} & &\\
        \cline{1-3}
        \tablehead{Duration of an episode} &
        \tablehead{Value-based vs.\ policy-based vs\ actor-critic (see \Cref{tab:value_based_vs_policy_based})} &
        
        \tablehead{Distribu-tional (see \Cref{tab:distributional})} &
        
        \tablehead{``then ...'' (method property)} & 
        \tablehead{``because ...'' (reason for method property)} \\
        \hline   

    \end{tabularx}

    \begin{tabularx}{\textwidth}{|p{0.07\textwidth}|p{0.1\textwidth}|p{0.08\textwidth}|p{0.117\textwidth}|p{0.5\textwidth}|}

    Short &
    Policy-based or actor-critic &
    No &
    Two options (with equal priority):
        \begin{titemize}
            \item Policy-based
            \item Actor-critic
        \end{titemize}
    &
        \begin{titemize}
            \item In theory, policy-based methods without a learned critic are suited for such situations since:
                \begin{titemize}
                    \item Updating the policy parameters only at the end of an episode is no problem if episodes are short.
                    \item One could profit from unbiased gradient estimates.
                \end{titemize}
            \item Actor-critic methods are also suited for such situations since:
                \begin{titemize}
                    \item Short episodes are no problem (but also no benefit) for actor-critic methods.
                    \item Even though policy-gradients may be biased in actor-critic algorithms, one could profit from lower variance in the gradient estimates compared to policy-based methods without a learned critic \cite[Section~13.8]{sutton2018}.
                \end{titemize}
            \item We conclude that the decision between a policy-based and an actor-critic algorithm is largely determined by the difference in bias and variance in the respective policy-gradient estimates. Which property (no bias or lower variance) weighs more can probably not be assessed in advance and experiments with both kinds of algorithms might be necessary.
        \end{titemize}
        \\ \hline
    
    Short &
    Policy-based or actor-critic &
    Yes &
    Actor-critic &
    \begin{titemize}
        \item See the row above. However, pure policy-based methods are typically not combined with distributional approaches since distributional approaches are based on a learned value function, which policy-based methods do not comprise.    
    \end{titemize}
    
    \\ \hline

    Long trajectories or continuing tasks & 
    Policy-based or actor-critic &
    Any &
    Actor-critic & 
        \begin{titemize}
            \item Policy-based methods without a learned critic are probably not suited for such situations since updating parameters only after completed episodes is often inefficient if episodes are long and even impossible for online learning in continuing tasks.
            \item Actor-critic methods are suited better for such situations since updates of the policy parameters are also possible during an ongoing episode. This is, in particular, desirable if episodes take a long time or are continuing.
        \end{titemize}
        \\ \hline
  
    \end{tabularx}
 
\end{table}
{\footnotesize
\begin{longtable}{|p{0.06\textwidth}|p{0.06\textwidth}|p{0.13\textwidth}|p{0.15\textwidth}|p{0.47\textwidth}|}
 
    \caption{Considerations for the decision between different variants of value-based RL algorithms (for a preceding decision between policy-based, actor-critic, and value-based algorithms, see \Cref{tab:value_based_vs_policy_based}). Related terms are explained in \Cref{sec:value_based_vs_policy_based}.
     }
    \label{tab:value_based_variants}
    \\
    \hline
    
    \multicolumn{3}{|p{0.25\textwidth}|}{\tablehead{``If ...'' (situation)}} & &\\
    \cline{1-3}
    \multicolumn{2}{|p{0.12\textwidth}|}{\tablehead{Environment properties}} & 
    \tablehead{Method properties} & &\\
    \cline{1-3}
    \multicolumn{1}{|b{0.06\textwidth}|}{\tablehead{State-space size}} &
    \multicolumn{1}{b{0.06\textwidth}|}{\tablehead{Action-space size}} &
    \multicolumn{1}{b{0.13\textwidth}|}{\tablehead{Value-based vs.\ policy-based vs.\ actor-critic (see \Cref{tab:value_based_vs_policy_based}) }} &
    \multicolumn{1}{b{0.15\textwidth}|}{\tablehead{``then ...'' (method property)}} & 
    \multicolumn{1}{b{0.47\textwidth}|}{\tablehead{``because ...'' (reason for method property)}} \\
    \hline

    \endfirsthead
      
    \hline
    \multicolumn{3}{|p{0.25\textwidth}|}{\tablehead{``If ...'' (situation)}} & &\\
    \cline{1-3}
    \multicolumn{2}{|p{0.12\textwidth}|}{\tablehead{Environment properties}} & 
    \tablehead{Method properties} & &\\
    \cline{1-3}
    \multicolumn{1}{|b{0.06\textwidth}|}{\tablehead{State-space size}} &
    \multicolumn{1}{b{0.06\textwidth}|}{\tablehead{Action-space size}} &
    \multicolumn{1}{b{0.13\textwidth}|}{\tablehead{Value-based vs.\ policy-based vs.\ actor-critic (see \Cref{tab:value_based_vs_policy_based}) }} &
    \multicolumn{1}{b{0.15\textwidth}|}{\tablehead{``then ...'' (method property)}} & 
    \multicolumn{1}{b{0.47\textwidth}|}{\tablehead{``because ...'' (reason for method property)}} \\
    \hline

    \endhead

        \multicolumn{2}{|p{0.15\textwidth}|}{Discrete and very small (as an indication, the Q-function, i.e.\ a table of shape $\left\lvert \mathcal{S} \right\rvert \times \left\lvert \mathcal{A} \right\rvert$ in this case, should easily fit into some memory that has acceptable access speeds)} &  
        Value-based &
        Tabular value-based &
        \begin{titemize}
        \item For very small state and action spaces, the action-value function $Q(s, a)$ could be tabular. For tabular action-value-based methods (in contrast to non-tabular methods), very good convergence guarantees exist \cite{watkins1989}, \cite[Chapters~5,6]{sutton2018}. This makes them the best choice in such situations. If state and action spaces are not very small (all rows below), tabular methods are usually not applicable and are therefore not considered in those cases.    
        \end{titemize}
        \\ \hline
        
        Any &  
        Discrete and small & 
        Value-based &
        Two options (priority according to the given order):
            \begin{titemize}
                \item Non-tabular value-based with exact maximization
                \item Non-tabular value-based with approximate maximization
            \end{titemize}
         &
            \begin{titemize}
                \item Value-based methods with exact maximization over actions are suited best for such situations since:
                \begin{titemize}
                    \item By using an appropriate function approximator that represents the action-value function, i.e.\ a sufficiently expressive neural-network architecture, any state-space sizes can be handled.
                    \item An exact maximization over actions is tractable if the number of actions is small.
                \end{titemize}
                \item Value-based methods with approximate maximization over actions are also suited for such situations but probably receive lower priority since there are no known advantages of using an approximate maximization in such situations that would be worth the computational overhead created by the approximating procedure.
            \end{titemize}
        \\ \hline
        
        Any &  
        Large discrete or low-dimen-sional continuous & 
        Value-based &
        Two options (priority according to the given order):
            \begin{titemize}
                \item Non-tabular value-based with approximate maximization and fixed search procedure
                \item Non-tabular value-based with approximate maximization and learned search procedure
            \end{titemize}
         &
        
            \begin{titemize}
                \item Value-based methods with exact maximization over actions are not suited for such situations because in large discrete or continuous action spaces, computing the argmax exactly is often hard. The reason is that $Q(s,a)$ is, in general, not concave and, in general, cannot be computed in closed form. 
                \item Value-based methods with approximate maximization over actions and fixed search procedure are suited for such situations since they are known to work well in large discrete or low-dimensional continuous action spaces \cite{kalashnikov2018, quillen2018}.
                \item Value-based methods with approximate maximization over actions and learned search procedure are also suited for such situations but probably receive lower priority since learning an additional function for the search procedure might add overhead to the training procedure without leading to better results in such situations \cite{vandewiele2020}.
            \end{titemize}
         \\ \hline

         \multicolumn{5}{|c|}{\centering $\downarrow$} \\ \hline
        
        Any &  
        Very large discrete or high-dimen-sional continuous & 
        Value-based &
        Non-tabular value-based with approximate maximization and learned search procedure &
            \begin{titemize}
                \item Value-based methods with exact maximization over actions are probably not suited for such situations since, in large discrete or continuous action spaces, computing the argmax exactly is usually not tractable (cf.\ row above).
                \item Value-based methods with approximate maximization over actions and fixed search procedure are probably not suited for such situations due to a high consumption of computational resources \cite{duan2016} and/or bad performance in high action dimensions \cite{vandewiele2020}.
                \item We conclude that value-based methods with approximate maximization over actions and learned search procedure are best suited for such situations since they alleviate the drawbacks of the methods mentioned above.
            \end{titemize}
        \\ \hline

\end{longtable}
}

\subsection{Value-function learning} \label{sec:value_function_learning}

In \Cref{tab:value_function_learning}, we provide some considerations that might be helpful for choosing the sub-algorithm for value-function learning.

The choice of the sub-algorithm for value-function learning, i.e.\ the sub-algorithm responsible for estimating the state-value function $V(s)$, the action-value function $Q(a,s)$, or the advantage function $A(a,s) = Q(a,s) - V(s)$ is a key design choice in value-based and actor-critic methods. Intuitively, action-value and advantage functions provide an assessment of the quality of the current policy in terms of the \emph{future cumulative reward} $\sum_t r_t$ (also called \emph{return}), where $r_t$ is short for the reward at time $t$ that could be expected from following the current policy $\pi(a\mid s)$ after state $s_t$, i.e.\ from time $t$ onward.

As a reminder, the action-value function $Q(s, a)$ is defined as the expected value of cumulative rewards for state-action pairs, i.e.\ $Q(s, a) = \mathbb{E}(\sum_{\tau=t}^{\Tmax} \gamma^{\tau+1} r_{\tau+1} \mid  s_t = s, a_t = a)$, where $r_t$ is the reward at time $t$, $\Tmax$ is the maximal duration of an episode, $0 < \gamma < 1$ is the discount factor that ensures convergence of the series for $\Tmax \rightarrow \infty$, and $s_t$ and $a_t$ are the state and action at time $t$, respectively. The state-value function $V(s)$ is defined analogously with an additional summation over all possible actions $a_t$ (or an integral in the case of continuous actions).

These sub-algorithms fundamentally determine how the agent learns how to act in the environment since the policy is either directly inferred from the value function (value-based methods) or the parameter updates of a parametric policy rely on a value function (actor-critic methods). Even for policy-based methods without a learned critic, where the updates of the policy do not directly rely on a learned value function, a state-value function is usually involved in the form of a baseline to increase training stability (see \Cref{sec:value_based_vs_policy_based} and \Cref{sec:stability}). 

In the following, we provide a brief overview using a high-level classification of the respective sub-algorithms. A detailed consideration of the sub-algorithms is beyond the scope of this work. We refer to the literature cited in \Cref{tab:value_function_learning} for further details and to \cite{sutton2018} for an introduction to value-function learning. In addition, we would also like to refer to~\cite{andrychowicz2021} for a comparison of some estimators, namely $n$-step, GAE, and V-trace, in an on-policy algorithm applied in a continuous environment.

The classes of algorithms we consider are:
\begin{itemize}
    \item \textbf{Monte Carlo} (MC) algorithms learn the value functions from cumulative rewards observed during entire sampled trajectories. Therefore, the parameters of the value function can only be updated after completed trajectories. The sample mean of the cumulative rewards obtained from multiple sampled episodes is an unbiased estimator of the expectation of the cumulative reward \cite[Chapter 5]{sutton2018}. It can, therefore, be used for learning the value function in a supervised manner.
    
    \item \textbf{One-step bootstrapping} algorithms, often summarized under the term temporal-difference (TD) learning, follow a different approach. Instead of observing cumulative rewards of entire episodes, a bootstrapping target (in the sense of an ``output target'') for the value function estimator is created from the last observed reward and an approximation of the value function calculated with the current parameterization (e.g.\ the current neural network), hence bootstrapped. As a simple example, assume we want to learn the discounted state-value function $V$ for some fixed policy $\pi$. Then, a gradient-based TD update rule for the estimator $V_\phi$ is given by $\phi \leftarrow \phi + \alpha(r_{t+1} + \gamma V_\phi(s_{t+1}) - V_\phi(s_t)) \nabla_\phi V_\phi(s_t)$, where $\alpha$ is the learning rate and the transition $s_t \rightarrow s_{t+1}$ is due to an action $a \sim \pi$ \cite[Chapter 9]{sutton2018}. Some of the most popular RL algorithms like $Q$-learning~\cite{watkins1989} and SARSA~\cite{rummery1994} are based on the concept of temporal-difference learning.
    
    \item \textbf{Eligibility traces, $n$-step methods}, and related algorithms usually provide a trade-off between one-step bootstrapping and MC learning in the sense that they consider the reward of multiple steps together with a bootstrapped value as the learning objective. For instance, replacing the objective $r_{t+1} + \gamma V_\phi(s_{t+1})$ by $r_{t+1} + \gamma r_{t+2} + \gamma^2 V_\phi(s_{t+2})$ in the example of the previous bullet point would make it a 2-step method. In general, MC and TD methods are special cases of eligibility traces and $n$-step methods. That is MC and TD methods can usually be recovered for certain values of an additional hyperparameter (e.g., \ $n=\infty$ for MC and $n=1$ for TD in $n$-step methods) \cite[Chapter~12]{sutton2018}, \cite{munos2016,vanhasselt2021}. That way, the benefits of one-step bootstrapping and MC methods can often be combined in one algorithm. Examples of such methods are: 
        \begin{itemize}
            \item TD($n$) (see \cite[Chapter~7]{sutton2018}, for instance)
            \item TD($\lambda$)~\cite{sutton1988}
            \item LSTD-Q($\lambda$)~\cite{Boyan2002}
            \item Q($\lambda$)~\cite{watkins1989,peng1994}
            \item Sarsa($\lambda$) \cite{rummery1994}
            \item TB($\lambda$) \cite{precup2000}
            \item ET($\lambda$) \cite{vanhasselt2021}
            \item GAE($\lambda$) \cite{schulman2015b}
            \item Retrace($\lambda$)~\cite{munos2016}
            \item GTD($\lambda$) \cite{maei2011}
            \item V-trace($n$) \cite{espeholt2018}
        \end{itemize}
\end{itemize}

{\footnotesize

\begin{longtable}{|p{0.08\textwidth}|p{0.06\textwidth}|p{0.06\textwidth}|p{0.1\textwidth}|p{0.12\textwidth}|p{0.42\textwidth}|}
    \caption{Considerations for value-function learning, i.e.\ the sub-algorithm for learning a state-value function $V(s)$, an action-value function $Q(a,s)$, or an advantage function $A(a,s) = Q(a,s) - V(a,s)$. This sub-algorithm is only required if \Cref{tab:value_based_vs_policy_based,tab:value_based_variants,tab:policy_based_variants} indicate a value-based or actor-critic algorithm. The relevant method properties are described in \Cref{sec:value_function_learning}. The Markov assumption is \emph{violated} in a decision process if knowledge about multiple states is necessary in order to assess the correct value of a state or state-action pair when acting according to a fixed policy. Rewards are \emph{sparse} if they are only obtained in a few states. Rewards are \emph{dense} if a reward is obtained in every state. \deceptive{}. Often, the sub-algorithm for value-function learning can be exchanged/adapted easily; some works from \Cref{tab:algorithms} already used different options for their experiments (as indicated by `/` in \Cref{tab:algorithms}, e.g., `TD/TD(n)` in algorithm `DQfD`).
    }
    \label{tab:value_function_learning}\\
    \hline
    
    \multicolumn{4}{|l|}{\tablehead{``If ...'' (situation)}} & &\\
    \cline{1-4}
    \multicolumn{3}{|b{0.2\textwidth}|}{\tablehead{Environment properties}} & 
    \multicolumn{1}{b{0.1\textwidth}|}{\tablehead{Method properties}} & &\\
    \cline{1-4}
    \multicolumn{1}{|b{0.08\textwidth}|}{\tablehead{Markov assumption}} & 
    \multicolumn{1}{b{0.06\textwidth}|}{\tablehead{Re-wards}} &
    \multicolumn{1}{b{0.06\textwidth}|}{\tablehead{Dura-tion of an episode}} & 
    \multicolumn{1}{b{0.1\textwidth}|}{\tablehead{Value-based vs.\ policy-based vs.\ actor-critic (see \Cref{tab:value_based_vs_policy_based,tab:value_based_variants,tab:policy_based_variants})}} &
    \multicolumn{1}{b{0.12\textwidth}|}{\tablehead{``then ...'' (method property)}} & 
    \multicolumn{1}{b{0.42\textwidth}|}{\tablehead{``because ...'' (reason for method property)}} \\
    \hline    
    
    \endfirsthead 
    
    \hline
    
    \multicolumn{4}{|l|}{\tablehead{``If ...'' (situation)}} & &\\
    \cline{1-4}
    \multicolumn{3}{|b{0.2\textwidth}|}{\tablehead{Environment properties}} & 
    \multicolumn{1}{b{0.1\textwidth}|}{\tablehead{Method properties}} & &\\
    \cline{1-4}
    \multicolumn{1}{|b{0.08\textwidth}|}{\tablehead{Markov assumption}} & 
    \multicolumn{1}{b{0.06\textwidth}|}{\tablehead{Re-wards}} &
    \multicolumn{1}{b{0.06\textwidth}|}{\tablehead{Dura-tion of an episode}} & 
    \multicolumn{1}{b{0.1\textwidth}|}{\tablehead{Value-based vs.\ policy-based vs.\ actor-critic (see \Cref{tab:value_based_vs_policy_based,tab:value_based_variants,tab:policy_based_variants})}} &
    \multicolumn{1}{b{0.12\textwidth}|}{\tablehead{``then ...'' (method property)}} & 
    \multicolumn{1}{b{0.42\textwidth}|}{\tablehead{``because ...'' (reason for method property)}} \\
    \hline

    \endhead
    
    Markov assumption violated (i.e. the optimal assessment of the value of a certain state or action at the current state requires knowledge about past observations/states) & 
    Any & 
    Short & 
    Value-based or actor-critic &
    Multiple options (with equal priority):
        \begin{titemize}
            \item Monte Carlo (MC) learning
            \item Eligibility traces or $n$-step methods
        \end{titemize}
    &

        \begin{titemize}
            \item In theory, Monte Carlo learning is suited for such situations since:
                \begin{titemize}
                    \item It is robust to violations of the Markov assumption as parameter updates are based on information from entire episodes.
                    \item Waiting for parameter updates until the end of an episode is no problem if episodes are short.
                \end{titemize}
            \item Eligibility traces and $n$-step methods are probably suited equally well to MC methods since:
                \begin{titemize}
                    \item They are also robust to violations of the Markov assumption with parameters tuned ``towards MC''.
                    \item Short episode lengths do not have a known negative impact on the performance on these methods.
                \end{titemize}
            \item In theory, TD methods are less suited for such tasks since:
                \begin{titemize}
                    \item They are less robust to violations of the Markov assumption as parameter updates are only based on information about single environment steps.
                    \item The fact that episodes are short does not compensate for the above disadvantage.
                \end{titemize}
        \end{titemize}
    
    \\ \hline

    \multicolumn{6}{|c|}{\centering $\downarrow$} \\ \hline
            
    Markov assumption holds & 
    Any  & 
    Short & 
    Value-based or actor-critic &
    Multiple options (with the highest priority to the first bullet point, equal but less priority to the second and third bullet point):
        \begin{titemize}
            \item Eligibility traces or $n$-step methods
            \item TD methods
            \item Monte Carlo (MC) learning
        \end{titemize}
    &
    
        \begin{titemize}
            \item In theory, Monte Carlo learning is suited for such situations since:
                \begin{titemize}
                    \item It propagates rewards obtained during an episode to every state or state-action pair involved in the episode within one update.
                    \item Waiting for parameter updates until the end of an episode is no problem if episodes are short.
                \end{titemize}
            \item In theory, TD learning is also suited for such situations since:
                \begin{titemize}
                    \item In the case of dense rewards and a holding Markov assumption, the agent receives immediate informative feedback about its actions at every TD update (while such updates are less informative if one or both of these properties do not hold, cf.\ the last row of this table).
                    \item In the case of sparse rewards, TD learning might still work in short episodes. The reason is that, even though propagation of rewards obtained at the end of an episode to involved state or state-action pairs requires multiple updates, this drawback is probably not significant if episodes are short (not many steps to propagate back). Examples like \cite[Example~6.2]{sutton2018} show that they can even outperform MC methods on such tasks.
                \end{titemize}
            \item We conclude that eligibility traces might be the best choice since they can be tuned to a trade-off between MC and TD that works best for the task at hand.
        \end{titemize} 
    \\ \hline

    Holds & 
    Non-decep-tive and quite dense & 
    Long episodes or continuing tasks & 
    Value-based or actor-critic &
    Multiple options (with priority according to the given order):
        \begin{titemize}
            \item TD methods
            \item Eligibility traces or $n$-step methods
        \end{titemize} 
    &
        \begin{titemize}
            \item Monte Carlo learning is probably not suited for such situations since updating parameters only after completed episodes is often inefficient if episodes are long and even impossible for online learning in continuing tasks.
            
            \item TD learning is suited for such situations since: 
                \begin{titemize}
                    \item Information obtained from one environment step will likely be informative as the Markov assumption holds and rewards are quite dense.
                    \item Due to their bootstrapping nature, TD methods are well-suited for long episodes or continuing tasks.
                \end{titemize}
                
            \item Eligibility traces and $n$-step methods might get slightly lower priority than TD in this situation. Even though they can be made ``like TD'' by appropriate hyperparameters and therefore allow for the same reasoning as for TD above, they are usually more complex to implement or to debug.
        \end{titemize} 
     \\ \hline
    
    \multicolumn{6}{|c|}{\centering $\downarrow$} \\ \hline

    \multicolumn{2}{|p{0.14\textwidth}|}{Markov assumption violated or rewards sparse, deceptive, or unclear in advance} & 
    Long episodes or continuing tasks & 
    Value-based or actor-critic &
    Eligibility traces or $n$-step methods &
        \begin{titemize}
            \item Monte Carlo learning is probably not suited for such situations since 
                \begin{titemize}
                    \item Updating parameters only after completed episodes is often inefficient if episodes are long and even impossible for online learning in continuing tasks.
                \end{titemize}
            \item TD learning is probably not suited for such situations since 
                \begin{titemize}
                    \item They are not robust to violations of the Markov assumption as parameter updates are only based on information about single environment steps.
                    \item Most individual steps are uninformative in an environment where rewards are obtained in only a few states and episodes are long. 
                \end{titemize}
            \item We conclude that eligibility traces and $n$-step methods are the best choices in this situation. They can account for violations of the Markov assumption and sparse rewards by tuning them ``towards MC'' while still being applicable to long or continuing tasks. 
        \end{titemize}
    \\ \hline

\end{longtable}
}

\subsection{Entropy regularization} \label{sec:entropy}

In \Cref{tab:entropy}, we provide considerations about when to apply entropy regularization. In the following, we explain the main idea together with various realizations.

The well-known dilemma in reinforcement learning is to decide between exploration of new policies (find unexplored actions/states through randomness) and exploitation of already discovered, well-working policies (good performance). In general, this trade-off is addressed by a behavior policy that contains, by its definition, a certain degree of randomness. In practice, a common problem is to ``tune'' the randomness, i.e.\ the amount of noise, of the behavior policy in a way that accounts well for the exploration/exploitation trade-off. The concept of \emph{entropy regularization} or \emph{maximum-entropy RL} addresses this problem in that it incorporates the policy's degree of randomness (i.e.\ how diverse the distribution of proposed actions is) into its training objective. While the standard objective of the policy is to maximize the expected future cumulative reward, the policy's objective in maximum-entropy RL is to maximize the reward \emph{plus} an entropy term (representing the policy's randomness). Furthermore, the entropy objective often prevents early, potentially sub-optimal convergence of the modes of the action distribution~\cite{haarnoja2017}. 

For maximum-entropy RL (in the form of the options listed in the bullet list below) to be applicable, the policy (which can be the `target policy' in policy-based and actor-critic algorithms, or a `proposal function' in value-based algorithms; for the sake of simplicity, we refer to both of these options with `policy` here) needs to be parameterized explicitly (not implicitly as in standard $Q$-learning). We refer to the parameters making up the policy, e.g.\ the weights of a neural network that parameterizes the action distribution, as `policy parameters`.

In practice, there exist several ways towards maximum-entropy RL; we list the options to choose from in the following.
\begin{itemize}
     \item A \textbf{per-state entropy regularization} term in the objective function of the policy can be used \cite{williams1991, mnih2016}. More precisely, a term of the form $\mathcal{H}(\pi_\theta({}\cdot{} \mid s_t))$ can be added to the reward in order to enhance the information entropy $\mathcal{H}$ of the policy in the current state $s_t \in \mathcal{S}$.
    
    \item While the regularization term discussed in the previous bullet point enhances the entropy of the policy at the current state $s_t$, \textbf{soft $Q$-learning} \cite{haarnoja2017, nachum2017} aims at enhancing the entropy of entire policy trajectories, i.e.\ $\sum_t \mathcal{H}(\pi_\theta({}\cdot{} \mid s_t))$, by considering the \emph{expected future cumulative entropy} in addition to the observed rewards. 
    
    \item A \textbf{Kullback--Leibler divergence} regularization term in the objective that is minimized with respect to policy parameters, i.e.\ a term $D_{\mathrm{KL}}(\pi_\theta({}\cdot{} \mid s_t) \parallel q({}\cdot{}))$, where $q({}\cdot{})$ can be any probability distribution and $D_{\mathrm{KL}}$ denotes the Kullback--Leibler divergence between two probability distributions. In practice, $q({}\cdot{})$ can be chosen to be the old action distribution \cite{schulman2017}, i.e.\ the policy before updating the parameters according to the objective. This can be interpreted as a relative entropy constraint on the parameter update because strong changes in the entropy of the stochastic policy are penalized and policies are usually initialized to have high entropy at the beginning of training (potentially drifting towards low entropy after some iterations).
    
    \item \textbf{Mutual information (MI)} between states and latent variables can also be used as an entropy regularizer. In~\cite{florensa2017}, maximization of the MI yields a diverse set of skills in a hierarchical policy. More precisely, in this case, the policy $\pi_\theta({}\cdot{} \mid s_t, c) $ depends on the current state $s_t$ and a hierarchically higher skill (see \Cref{sec:hierarchical}), represented as latent variable $c$. While $\mathcal{H}(\pi_\theta({}\cdot{} \mid s_t) )$ is intractable due to the integration over $c$, the MI given by $\mathcal{H}(c) - \mathcal{H}(c \mid s_t)$ can be computed. Note that the goal here is to have a unique action associated with a certain latent code (the entropy $\mathcal{H}(c \mid s_t)$ needs to be reduced to this end), which is different from the goal of learning a diverse set of actions described in the above bullet points; hence, the application of MI regularization to a specific problem probably requires a very detailed examination of the policy (and potentially involved latent variables).
    
\end{itemize}

\begin{table}
    \centering
    \footnotesize
    \caption{
    Considerations for enhancing the entropy of a stochastic policy. This is only possible for policy-based and actor-critic methods or for value-based methods with a learned search procedure as explained in \Cref{sec:entropy}. Rewards are \emph{sparse} if they are only obtained in a few states. Rewards are \emph{dense} if a reward is obtained in every state. \deceptive{}}
    \label{tab:entropy}
    \begin{tabularx}{\textwidth}{|b{0.1\textwidth}|b{0.1\textwidth}|b{0.1\textwidth}|b{0.17\textwidth}|b{0.4\textwidth}|}
    
        \hline
        
        \multicolumn{3}{|b{0.3\textwidth}|}{\tablehead{``If ...'' (situation)}} & &\\
        \cline{1-3}
        \tablehead{Environ-ment properties} & 
        \multicolumn{2}{b{0.2\textwidth}|}{\tablehead{Method properties}} & &\\
        \cline{1-3}
        \tablehead{Rewards} &
        \tablehead{Value-based vs.\ policy-based vs.\ actor-critic (see \Cref{tab:value_based_vs_policy_based,tab:value_based_variants,tab:policy_based_variants})} &
        \tablehead{Target policy (see \Cref{tab:stochastic_vs_deterministic})} &
        \tablehead{``then ...'' (method property)} & 
        \tablehead{``because ...'' (reason for method property)} \\
        \hline    

        \end{tabularx}

        \begin{tabularx}{\textwidth}{|p{0.1\textwidth}|p{0.1\textwidth}|p{0.1\textwidth}|p{0.17\textwidth}|p{0.4\textwidth}|}
 
        Quite dense and non-deceptive & 
        \multicolumn{2}{p{0.2265\textwidth}|}{
            \begin{titemize}
                \item Policy-based and stochastic target policy
                \item Actor-critic and stochastic target policy 
                \item Value-based with learned search procedure (the search procedure should be stochastic, i.e.\ the proposal distribution, cf.\ \Cref{sec:value_based_policies}, should not be a deterministic function)
            \end{titemize}

        }
        &
            Low priority (in the sense of ``the decision has lower priority than other decisions''): 
            \begin{titemize}
                \item Soft Q-learning (consider first)
                \item or KL regularization
                \item or Per-state entropy regularization
                \item or Mutual information regularization
            \end{titemize}
        &
        
            \begin{titemize}
                 \item Enhanced entropy might prevent a stochastic policy from premature convergence but requires an additional hyperparameter to tune (often a temperature parameter that assigns importance to the entropy term in the loss function). 
                 \item Among the options to enhance entropy, soft Q-learning seems to be the most popular, and it is powerful as it considers the entropy of the distribution over entire trajectories (not only locally at individual states as in per-state entropy regularization, for instance)
                 \item It is not always straightforward to compute the entropy of some policies, e.g.\ for Gaussian mixture \cite{baram2021} or black-box policies \cite{tang2018}, which could make it lower priority due to additional effort required (in terms of implementation, computation).
                 \item A better exploration resulting from enhanced entropy might not be crucial if the agent receives frequent and clear feedback on his actions from quite dense and non-deceptive rewards. 
             \end{titemize}
       \\ \hline
        
        Sparse, deceptive, or varying (e.g., \ stochastic rewards) 
        & 
        \multicolumn{2}{p{0.2265\textwidth}|}{ 
            \begin{titemize}
                \item Policy-based and stochastic target policy
                \item Actor-critic and stochastic target policy 
                \item Value-based with learned search procedure (the search procedure should be stochastic, i.e.\ the proposal distribution, cf.\ \Cref{sec:value_based_policies}, should not be a deterministic function)
            \end{titemize}
        }
        &
        Multiple options (with priority according to the given order):
        \begin{titemize}
            \item Soft Q-learning
            \item KL regularization
            \item Per-state entropy regularization
            \item Mutual information regularization
        \end{titemize}
         &
        
        \begin{titemize}
             \item Enhanced entropy ensures good exploration \cite{haarnoja2018}, which helps in sparse- and/or deceptive-reward settings. 
             \item High entropy is known to increase robustness to dynamic environments, which might include a varying reward function \cite{eysenbach2022maximum}. 
             \item If computing the entropy is impossible, e.g.\ due to a non-invertible stochastic process, regularization can potentially be based on a mutual information objective. To examine whether this is possible, \cite{florensa2017} might be an initial reference. See the row above for the order of different ways of entropy enhancement.
        \end{titemize}
        
        \\ \hline
        
    \end{tabularx}
 
\end{table}

\section{Action-distribution families} \label{sec:policies}

The goal of RL is to find an optimal policy $\pi(a\mid s)$ that maps state $s$ to a probability for action $a$ in a certain environment. In this section, we discuss which \emph{action-distribution family} to use in the RL algorithm selected from \Cref{tab:algorithms}. While the policy defines a specific behavior of the agent (which is optimized during training), the action-distribution family is the ``blueprint'' for the policy and needs to be chosen a priori. Mathematically, the action-distribution family is a family of probability distributions over actions, often characterized by a certain shape but without specific parameters (e.g.\ a Gaussian without having its mean and variance parameters specified). The goal of the RL training process is to find these parameters, i.e.\ to select/learn the optimal policy from the action-distribution family in the given environment. Hence, the choice of the action-distribution family is important as it pre-defines the possible behavior of the agent during training (behavior-action-distribution family) and the possible result of the training process (target-action-distribution family).

\subsection{Value-based action-distribution families (action-distribution families for value-based algorithms)} \label{sec:value_based_policies}

\begin{table}
    \centering
    \footnotesize
   \caption{
   Overview of value-based action-distribution families. This table is relevant if \Cref{tab:value_based_vs_policy_based} indicates a value-based algorithm. If the selected algorithm is further on-policy, cf.\ \Cref{tab:on_policy_vs_off_policy} (on-policy vs.\ off-policy), one choice needs to be made (behavior-action-distribution family and target-action-distribution family are equal). If the selected algorithm is off-policy, different choices for the behavior-action-distribution family and target-action-distribution family are possible from this table. An optimal choice could require empirical evaluation but we provide some guidance in the respective column. Possible NN architectures for the $Q$-function, on which all of the policies in this table rely, are discussed in \Cref{sec:architectures}. 
   }
    \label{tab:value_based_policies}
    \begin{tabularx}{\textwidth}{|p{0.1\textwidth}|p{0.2\textwidth}|p{0.2\textwidth}|p{0.395\textwidth}|}
    \hline
        \tablehead{Action-distribution family} & 
        \tablehead{Hyperparameters and effects} & 
        \tablehead{Stochastic/deter\-min\-istic (see \Cref{tab:stochastic_vs_deterministic})} &
        \tablehead{Guidance for selection}
        \\ \hline

    Greedy &
    None &
    Stochastic if $Q$ and/or the $\argmax$ computation are stochastic; deterministic otherwise &
    \begin{titemize}
        \item Typical deterministic target-action-distribution family in value-based off-policy algorithms
        \item Rarely used as a behavior-action-distribution family since exploration might be limited
    \end{titemize}
    \\
    \hline
    
    $\varepsilon$-greedy~\cite{mnih2015} &
    Larger $\varepsilon$ leads to more exploration and less exploitation &
    Stochastic &
    The ``standard'' value-based behavior-action-distribution family (and ``standard'' target-action-distribution family in on-policy value-based algorithms)
    \\ \hline
    
    Boltzmann exploration &
    None &
    Stochastic &
    Usually only applicable in discrete environments where the $Q$-value of each possible action can be computed.
    \\ \hline

    Randomized value functions~\cite{osband2016} &
    Larger $K$ leads to faster learning &
    Stochastic &
    Initially tested only on toy experiments, see this paper for more details:~\cite{osband2019}
    \\
    \hline
    
    Noisy nets~\cite{fortunato2018} & 
    The choice of the noise distribution(s) as well as the distribution parameters (fixed noise) or the initialisation of the distribution parameters (learned noise) are hyperparameters in this policy; see \cite{fortunato2018} for an analysis of Gaussian noise &
    Stochastic &
    \begin{titemize}
        \item Can be used instead of the $\varepsilon$-greedy action-distribution family
        \item Has shown to be beneficial for game playing~\cite{fortunato2018} but it is unclear whether it is better than $\varepsilon$-greedy in general
    \end{titemize}
    \\
    \hline

    \end{tabularx}
\end{table}

If \Cref{tab:value_based_vs_policy_based} indicates a value-based algorithm, a value-based action-distribution family needs to be chosen. We list options for value-based action-distribution families in \Cref{tab:value_based_policies} and explain them in the following. We further describe how they can be parameterized by a NN. Recall from \Cref{sec:value_based_vs_policy_based} that the action-value function estimates the expected future cumulative reward, i.e.\ $Q(s,a) = \mathbb{E}(\sum_{\tau=t}^{\Tmax}r_{\tau+1} \mid  s_t = s, a_t = a)$.

\begin{itemize}
    \item The \textbf{greedy} value-based action-distribution family comprises all policies that maximize an action-value function in a certain state, i.e.\ $a_t = \argmax_a Q(s_t, a)$. The computation of the $\argmax$ can be exact or approximated according to the choice made in \Cref{tab:value_based_variants} (value-based variants).
    \item In the \textbf{$\varepsilon$-greedy} value-based action-distribution family, a uniformly sampled random action is chosen by probability $\varepsilon$, and with probability $1-\varepsilon$ the action given by $a_t = \argmax_a Q(s_t, a)$ is chosen.
    \item \textbf{Boltzmann exploration} denotes an action-distribution family in which the probability of each action is given by the softmax over the $Q$-values of the action.
    \item \textbf{Randomized value functions} denote an action-distribution family in which the agent randomly chooses among $K$ policies inferred from $K$ different action-value functions, i.e. $a_t = \argmax_a Q_k (s_t, a)$, where $k$ is sampled uniformly from $\{1, \ldots, K\}$.
    \item \textbf{Noisy nets} denote an action-distribution family in which $Q$ is a NN with stochastic weights (noise added to the parameters of the NN).
\end{itemize}

\paragraph{Parameterization with a NN}
Depending on the kind of non-tabular value-based RL algorithm, different parts of the policy are typically parameterized by a NN: 
\begin{itemize}
    \item In \textbf{value-based RL with exact maximization} and \textbf{value-based RL with approximate maximization and fixed search procedure}, the $Q$-function is parameterized by a NN $Q_\theta$.
    \item In \textbf{value-based RL with approximate maximization and learned search procedure}, the $Q$-function and a so-called proposal distribution are represented by NNs. The proposal distribution defines a probability distribution over actions (similar to a stochastic actor in actor-critic methods). It is used to sample a small number of actions to make the computation of $\argmax_a Q_\phi(a,s)$ tractable in large action spaces.
\end{itemize}
For considerations about NN architectures in RL, see \Cref{sec:architectures}.


{\footnotesize
    \begin{longtable}{|p{0.19\textwidth}|p{0.12\textwidth}|p{0.15\textwidth}|p{0.12\textwidth}|p{0.28\textwidth}|}
    
    \caption{
    Overview of parametric action-distribution families. A choice from this table is necessary if \Cref{tab:value_based_vs_policy_based} indicates a policy-based or actor-critic algorithm. The choice should be made based on the last four columns of this table, i.e.\ based on the decision between a stochastic and deterministic policy, the expressiveness of the distribution as indicated by \Cref{tab:expressiveness}, the action space, and potential further considerations (summarized in the ``Comments'' column). If the selected RL algorithm is on-policy, cf.\ \Cref{tab:on_policy_vs_off_policy} (on-policy vs.\ off-policy), one choice needs to be made (behavior-action-distribution family and target-action-distribution family are equal). If the selected algorithm is off-policy, different choices for the behavior-action-distribution family and target-action-distribution family are possible from this table. Neural-network architectures for policy parameterization are discussed in \Cref{sec:architectures}. All of the parametric action-distribution families in this table allow for so-called implicit reparameterization gradients, which allow for backpropagation through the sampling process, see related literature for details~\cite{figurnov2018, jang2017, kingma2014, rezende14}.
    } 
    \label{tab:stochastic_parametric_policies}\\
    
    \hline
         & 
        \multicolumn{4}{l|}{\tablehead{Guidance for selection}} \\ \hline
        \tablehead{Action-distribution family} &
        \tablehead{Stochastic vs.\ deterministic (see \Cref{tab:stochastic_vs_deterministic})} &
        \tablehead{Potential expres\-sive\-ness of the distribution (see \Cref{tab:expressiveness})} &
        \tablehead{Action space} &
        \tablehead{Comments}
        \\ \hline
    
    \endfirsthead 
        
    \hline
      \tablehead{Action-distribution family} & 
        \multicolumn{4}{l}{\tablehead{Guidance for selection}} \\ \hline
        &
        \tablehead{Stochastic vs.\ deterministic (see \Cref{tab:stochastic_vs_deterministic})} &
        \tablehead{Potential expres\-sive\-ness of the distribution (see \Cref{tab:expressiveness})} &
        \tablehead{Action space} &
        \tablehead{Comments}
        \\ \hline
    \endhead
        
        Categorical distribution, see e.g.~\cite{abdolmaleki2018, wang2017, cesabianchi2017} &
        Stochastic &
        The more bins, the higher expressiveness &
        Discrete &
        The standard parametric action-distribution family for discrete action spaces
        \\ \hline
        
        Gaussian distribution, see e.g.\ \cite{schulman2017, silver2014} &
    
        Stochastic &
        Low &
        Con\-tin\-u\-ous &
    
        Often considered a ``standard'' stochastic policy for continuous actions if there is no need for high expressiveness
        
        \\ \hline
  
        Gaussian mixture distribution, see e.g.\ \cite{baram2021} & 
        Stochastic &
        High (depending on the number of mixture components) &
        Con\-tin\-u\-ous &
        Probably the first choice if a Gaussian is not expressive enough
        \\ \hline
        
        Normalizing flows, see e.g.\ \cite{ward2019, mazoure2020, tang2018} & 

        Stochastic &
        High \newline (depending on the flow parameterization) &
        Con\-tin\-u\-ous &
        Allows for complex policies (in the sense of many modes) and closed-form computation of probability density
        \\ \hline
        
        Stochastic networks and black-box policies, see e.g.\ \cite{fortunato2018, tang2018} (can represent an arbitrary probability distribution) & 
        Stochastic &
        High \newline (depending on the NN architecture) \newline &
        Any  &
        Arbitrarily complex behavior can be modeled (arbitrarily many modes), probably not suited if probability densities need to be computed in closed form, e.g.\ for computation of entropy (cf.\ \Cref{sec:entropy})
        
        \\ \hline
        
        Any continuous distribution followed by $\tanh$,
        see e.g.\ \cite{haarnoja2018} & 
        Stochastic &
        Depends on the used continuous distribution &
        Con\-tin\-u\-ous, bounded &
        Standard approach for bounded action spaces
        \\ \hline
        
        Beta distribution, see e.g.\ \cite{chou2017} & 
        Stochastic &
        Low &
        Con\-tin\-u\-ous, bounded &
        Can be considered as an alternative to tanh if bounded actions are required
        \\ \hline
   
        Deterministic &
        Deterministic &
        Very low (not a probability distribution) &
        Continuous &
        Usually not used in on-policy algorithms as exploration is very limited
        \\ \hline

        Added noise &
        Stochastic &
        Depends on the noise distribution but typically low &
        Continuous &
        Often used as the behavior-action-distribution family if the RL algorithm is off-policy and the target-action-distribution family is deterministic, usually not used in on-policy algorithms
        
            \\ \hline
            
    \end{longtable}
}

\subsection{Parametric action-distribution families (action-distribution families for policy-based and actor-critic algorithms)} \label{sec:parametric_policies}

If \Cref{tab:value_based_vs_policy_based} indicates a policy-based or actor-critic algorithm, a parametric action-distribution family needs to be chosen.
We list options for parametric action-distribution families in \Cref{tab:stochastic_parametric_policies}. We explain these options, their parameterization with a NN, and related terms in the following. Recall from \Cref{sec:value_based_vs_policy_based} that, in policy-based and actor-critic algorithms, the policy $\pi_\theta$ is parameterized directly with parameters $\theta$ and not inferred from an action-value function (as in value-based RL, see \Cref{sec:value_based_policies}). 

\begin{itemize}
    \item The \textbf{categorical distribution} provides probabilities for a discrete set of actions. The NN maps observations to the logits of the action bins. The distribution can be arbitrarily factorized, i.e.\ assuming conditional independence between arbitrary (subsets of) actions.
    \item The \textbf{Gaussian distribution} is continuous with a single mode. The NN maps observations to the mean and covariance matrix of the Gaussian. The covariance matrix can be restricted (e.g. full, diagonal, state-independent).
    \item The \textbf{Gaussian mixture distribution} is continuous with an arbitrary number of modes (depending on the number of Gaussians that are mixed). The NN maps observations to the means, covariance matrices, and weights of multiple Gaussians. The covariance matrices can be restricted (e.g. full, diagonal, state-independent).
    \item \textbf{Normalizing flows} can represent complex continuous probability distributions with an arbitrary number of modes. The first layers of the NN map observations to the parameters of a simple base distribution (e.g. Gaussian), which is then transformed by a normalizing flow. Many variants exist, e.g.\ Real-NVP or radial flows.
    \item \textbf{Stochastic networks} and \textbf{black-box} distributions can represent an arbitrary probability distribution. The NN maps observations stochastically to actions, e.g.\ by using stochastic weights or random noise as an additional input. Many variants can be created by varying the transformation, i.e.\ by changing the NN architecture that represents this transformation. However, it can be difficult to compute the probability density in closed form.
    \item The \textbf{tanh} function can be used to incorporate boundaries into continuous action-distribution families.
    \item The \textbf{beta distribution} is a bounded probability distribution. The NN maps observations to the parameters of the beta distribution. One of the parameters could be fixed or learned directly rather than output by the NN.
    \item The \textbf{deterministic} action-distribution family comprises all policies that assign a probability of one to exactly one action. The NN maps observations to a single action.
    \item action-distribution families with \textbf{added noise} rely on a deterministic prediction to which they add noise. Compared to other stochastic parametric action-distribution families, the parameters of the noise distribution are usually chosen manually and are not learnable. Popular choices for the distribution of the sampled noise are the Ornstein--Uhlenbeck process~\cite{uhlenbeck1930} (yielding temporally correlated noise) or Gaussian distributions (yielding non-correlated noise).
\end{itemize}

For considerations about NN architectures in RL, see \Cref{sec:architectures}.

\paragraph{Expressiveness of action distributions}

In \Cref{tab:expressiveness}, we provide considerations about the required expressiveness of the action distributions, i.e.\ the distribution from which an action is sampled at a certain state. In the following, we explain this term together with various examples.

The action distributions modeled by stochastic parametric policies can be distinguished according to their expressiveness. For instance, a normalizing flow can output a complex action distribution with several modes, whereas a NN whose outputs are used as the parameters of a Gaussian distribution can produce only unimodal Gaussian action distributions. The complexity of the probability distribution of actions is not to be confused with the complexity of the mapping from a state to the parameters of the distribution.\footnote{Nor with other reasons for the complexity of state-space trajectories. For example, simple sequences of folding and cutting paper or choosing directions in a maze might yield complex state-space trajectories.}
 
We distinguish two kinds of action-distribution families according to their expressiveness:
\begin{itemize}
        \item \textbf{Simple action-distribution families} (with \textbf{low expressiveness}) are families of distributions with only a few possible modes like the Gaussian or beta distribution. Also, categorical distributions with only a few bins fall into this category. In contrast to expressive action-distribution families (see below), the complexity of the action distribution is limited by the simple shape of such distributions. 
        
        \item \textbf{Expressive action-distribution families} are,  for  example,  a  Gaussian  mixture  distribution \cite{baram2021} with a sufficient number of mixture components,  normalizing flows \cite{ward2019, mazoure2020, tang2018}, or non-invertible black-box policies \cite{tang2018} for continuous actions, but also categorical distributions in large discrete action spaces. Those distribution families allow for multiple modes in the action distribution and, therefore, for modeling advanced stochastic behavior.
\end{itemize}

\begin{table}
    \centering
    \footnotesize
    \caption{Considerations for learning a stochastic policy with high or low complexity of action distributions (given a state). These terms are introduced in \Cref{sec:parametric_policies}. Rewards are \emph{sparse} if they are only obtained in a few states. Rewards are \emph{dense} if a reward is obtained in every state. \deceptive{}}
    \label{tab:expressiveness}
    \begin{tabularx}{\textwidth}{|b{0.22\textwidth}|b{0.1\textwidth}|b{0.1\textwidth}|b{0.15\textwidth}|b{0.3\textwidth}|}
        \hline
        
        \multicolumn{3}{|b{0.42\textwidth}|}{\tablehead{``If ...'' (situation)}} & &\\
        \cline{1-3}
        \tablehead{Environment properties} & 
        \multicolumn{2}{b{0.2\textwidth}|}{\tablehead{Method properties}} & &\\
        \cline{1-3}
        \tablehead{Rewards and optimal behavior} &
        \tablehead{Value-based vs.\ policy-based vs.\ actor-critic (see \Cref{tab:value_based_vs_policy_based})} &
        \tablehead{Stochastic vs.\ deterministic target policy (see \Cref{tab:stochastic_vs_deterministic})} &
        \tablehead{``then ...'' (policy property)} & 
        \tablehead{``because ...'' (reason for policy property)} \\
        \hline    

        \end{tabularx}
        \begin{tabularx}{\textwidth}{|p{0.22\textwidth}|p{0.1\textwidth}|p{0.1\textwidth}|p{0.15\textwidth}|p{0.3\textwidth}|}
 
        Rewards quite dense and non-deceptive, (near-) optimal behavior with at most a few modes & 
        Policy-based or actor-critic &
        Stochastic target policy &
        Two options (with priority according to the given order):
            \begin{titemize}
                 \item Action distribution with low expressiveness (e.g.\ Gaussian)
                 \item Action distribution with high expressiveness (see row below)
            \end{titemize}
        &
        
        \begin{titemize}
            \item Action distributions with low expressiveness are the default in the reinforcement learning literature for such situations. However, it could be worth trying a more expressive policy distribution for the sake of a potentially reduced number of parameters (see below).
            \item Simple action distributions can, at most, cover a few action modes. This is sufficient if it is expected to express (near-) optimal behavior.
        \end{titemize}
        \\ \hline
        
        Rewards sparse, deceptive, or unclear in advance, or (near-) optimal behavior expected to have multiple modes (e.g.\ in competitive games where a fixed strategy can be figured out and exploited by the opponent) &
        Policy-based or actor-critic &
        Stochastic target policy &
        Policy distribution with high expressiveness (e.g.\ Gaussian mixture, normalizing flows) &

            \begin{titemize}
            \item Expressive action distributions can lead to faster discovery of good actions in sparse- or deceptive-reward settings while, in addition, reducing the number of 
            network weights (i.e.\ being more memory efficient)
            \cite{mazoure2020,ward2019}. Therefore, they are better suited than less expressive action distributions in such situations. 
            \item Expressive action distributions can cover multiple action modes.
            \end{titemize}
        
        \\ \hline
        
    \end{tabularx}
   
\end{table}

\section{Final selection steps} \label{sec:final-steps}
If the selection of methods as per the previous sections still leaves us with more than one option to choose from, the following approaches can be recommended for making a final choice:
\begin{itemize}
    \item Run experiments with several of those methods.
    \item Try methods that perform well on similar problems, according to literature (for example literature that introduces recent methods and compares them to other methods, or literature that uses recent methods).
\end{itemize}

\section{Neural-network architectures} \label{sec:architectures}

In this section, we list several examples of neural network (NN) architectures that are frequently implemented in RL to approximate a policy $\pi$ and/or one or multiple value functions, i.e.\ $Q$, $V$, and/or $A$. In addition, we give a brief overview of situations, when those architectures are applied in existent work. Note that, in general, those functions do not necessarily need to be represented by an NN (they can also be tabular or linear functions etc.). Still, neural networks have shown to be best suited for complex and challenging tasks in recent years. Apart from architectural subtleties, the training of deep RL models follows common deep learning practices. Popular references in this regard include \cite{goodfellow2016} and \cite{chollet2021}.

\paragraph{Fully Connected NNs} Fully connected architectures are usually applied if the state space is rather small, e.g.\ a set of atomic measurements. Examples: \cite{schulman2015a, lillicrap2016}. 

\paragraph{Time-Recurrent NNs} Recurrent architectures are usually applied in partially observable Markov decision processes (MDPs) where the current observation does not contain all information about the current state. Time-recurrent NNs have also been shown to perform well for fully observable MDPs. Examples: \cite{kapturowski2019, espeholt2018, vandewiele2020, mnih2015}.

\paragraph{CNNs} Convolutional architectures and/or the attention mechanism are usually applied whenever state observations consist of Euclidean-grid data like images or video sequences, i.e.~when feature extraction should be local and equally good regardless of the location of features. Examples: \cite{mnih2015, mnih2016, schulman2015a, lillicrap2016}.

\paragraph{Dueling NNs} Dueling architectures are primarily used for the approximation of the $Q$-function. The idea is to split state- and action-dependent information of $Q(s, a)$ into branches to better generalize learning across actions \cite{wang2016}.

\paragraph{Parameter sharing} Some parameters and features can be shared among multiple networks, for example, among a policy and a value function network, in order to reduce the total number of parameters and, therefore, lower memory consumption. Examples: \cite{schulman2017, mnih2016, vandewiele2020}. However, it should be noted that the lower memory consumption may come at the cost of worse test results as outlined in~\cite{andrychowicz2021}.

\paragraph{Broadcasting} If state and/or action representations have a different nature, e.g.\ images, atomic values, or feature vectors, they must be merged. This can be done using broadcasting operations, see \cite{kalashnikov2018}, for instance.

\paragraph{Seq2Seq} Sequence-to-sequence architectures have been successfully applied to high-dimensional spaces where individual dimensions are discretized \cite{metz2017}.

\section{Training of neural networks in RL} \label{sec:training}
In this section, we provide a list of approaches that can be applied during NN training to tackle two typical objectives in RL: a stable training process and good test performance of a trained model.

\subsection{Improving stability of the learning process} \label{sec:stability}
A general problem in RL is that the learning process suffers from instability. This means that the performance of an agent is not increasing monotonically during training and different random seeds could lead to entirely different results \cite{henderson2018}.

\citet{sutton2018} identified three factors that increase the risk of unstable training, while the effect is most severe if they are all present at a time: The approximation of the action-value function $Q$ (as opposed to tabular approximation) (e.g.\ through a neural network $Q^\phi$ with trainable parameters $\phi$), bootstrapping (\Cref{sec:value_function_learning}), and off-policy training (\Cref{sec:definition-on-off-policy}). They call a combination of these factors the \emph{deadly triad} and argue that going for an on-policy method in order to remove at least one of the three factors is often a reasonable way to achieve more training stability \cite[Chapter 11]{sutton2018}.

Further approaches that have proven beneficial for more stability in the training process and which can be applied to many deep RL algorithms, often independent of the algorithm, include:

\begin{itemize}
    \item \textbf{Trust regions} \cite{schulman2015a, wu2017} that constrain the parameter updates and/or \textbf{entropy regularization} (KL regularization can be interpreted as a soft form of trust regions and Soft $Q$-learning seems to also have a positive impact on training stability~\cite{haarnoja2018})
    \item \textbf{Clipping parts of the objective functions} \cite{schulman2017}
    \item \textbf{Clipping the gradients} of the policy and/or value function parameters with respect to the objectives (e.g.\ cumulative reward, entropy)
    \item \textbf{Target networks} \cite{mnih2015, lillicrap2016, haarnoja2018}, i.e.\ updating parameters used for the computation of bootstrapping targets in one-step or n-step bootstrapping methods (see \Cref{sec:value_function_learning}) less frequently than the parameters that are updated using this objective
    \item \textbf{Implicit parameterization} of stochastic action distributions using the reparameterization trick~\cite{figurnov2018, jang2017}, e.g.\ used in \cite{haarnoja2018} for a Gaussian distribution
    \item Learned state- or time-dependent \textbf{baselines in policy objectives}, usually related to the cumulative reward \cite{williams1992, mnih2016, duan2016}
    \item \textbf{Weight averaging} during training~\cite{nikishin2018}
\end{itemize}

\subsection{Improving test results} \label{sec:improving_results}

If a working RL algorithm has been chosen for a certain task at hand, there exist several ways to squeeze the maximum out of the model. Of course, as in most machine learning fields, an RL algorithm has various parameters that can be tuned for the best performance. Apart from parameter tuning, we would like to mention some more techniques to improve final test performance independent of the algorithmic choices made before. Such techniques may require additional training time as well as implementation effort and include:
\begin{itemize}
    \item \textbf{Learning a diverse set of policies}: This technique has already been mentioned in \Cref{tab:distributed} in the context of distributed learning. However, even without a distributed architecture, it is possible to learn a diverse set of policies (e.g.\ from different action-distribution families, cf.\ \Cref{sec:policies}, or different parameterizations of a policy). Such an approach (together with an adaptive mechanism of switching between a family of policies) was part of a method outperforming the human baseline on all 57 Atari benchmarks for the first time \cite{badia2020}.
    
    \item \textbf{Intrinsic motivation}: In the standard RL setting, the agent receives extrinsic rewards based on the state of the external environment. However, there is also the option to equip the agent with some sort of intrinsic motivation (during training or as a form of pre-training), representing the analogous intrinsic behavioral drive in animals \cite{harlow1950, singh2010}. This may improve exploration and, therefore, test performance, particularly in environments in which it is difficult, or impossible, to shape an appropriate external reward function and for which also no expert system is available (cf.\ \Cref{sec:imitation_learning}). A  survey about techniques to incorporate the concept of intrinsic motivation into an RL algorithm can be found in \cite{aubret2019}. Further, several intrinsic approaches were compared in~\cite{laskin2021}.
    
    \item \textbf{Data augmentation}: While the concept of data augmentation is commonly used in domains like computer vision~\cite{shorten2019}, it seems to have attracted less attention in RL so far. Yet, data augmentation could be effective in model-free RL~\cite{yarats2021}.
    
    \item \textbf{Low-level design choices}: Low-level design choices in the implementation of a deep RL method (for example, the composition of mini-batches, initialization) can also have an impact on the final test performance of RL agents~\cite{andrychowicz2021, engstrom2020}. Even though the provided references focus on on-policy actor-critic methods in continuous environments, their findings and practical recommendations might also be valuable for other classes of RL algorithms and in alternative environments. Apart from the algorithmic implementation, there might also be the possibility to design (parts of) the environment, e.g.\ initial states or rewards, especially in simulation-based applications. An analysis of environmental design choices on the example of simulated locomotion can be found in~\cite{reda2020}.
\end{itemize}

\section{Discussion and conclusions}

From the sheer number of tables and algorithms in this overview paper it can be clearly deduced that, so far, there is no ``ultimate'' model-free RL algorithm working for any situation or across different environments equally well. Many of the algorithms in \Cref{tab:algorithms} improve over one another with respect to one or more of common RL issues, namely time and sample efficiency, test performance, explainability, safety, training stability, and reproducibility, but there is no single algorithm that addresses all of these challenges best. 

In general, approaches that try to unify several concepts in a single algorithm, e.g.\ bridging the gap between value- and policy-based algorithms \cite{schulman2017b} or between on- and off-policy algorithms \cite{nachum2017}, with the aspiration to unify the best of multiple worlds seem to be promising. Such methods have been able to overcome obstacles thought to be insurmountable in the past.

Selecting an appropriate RL algorithm for a certain environment is straightforward when the environment's properties are well-known and the presented considerations can be easily taken into account step by step. On the other hand, environmental properties themselves may be hard to evaluate in advance (e.g., whether training stability or sample efficiency is more important). In these cases, it could make sense to change the order among decisive steps and to conduct experiments in between. For example, one could first decide on a certain family of action distributions and then consider going for an on- or off-policy algorithm based on experiments with instances of both possibilities before making a final decision. We also note that many works test their methods on a few popular benchmark environments, which is arguably necessary to compare them in a structured way. Still, insights may not generalize to every situation. Therefore, we encourage practitioners to try new compositions of sub-routines and to always take advice with a grain of salt.

Finally, we are aware that this manuscript remains incomplete, even regarding the core class of model-free deep RL algorithms for control tasks and the respective concepts we try to survey. Also, we have not covered recent advances in the field, e.g.\ related to Transformers and large language models. Still, we hope to give RL practitioners a hand in overlooking the ever-faster growing field and to aid in assembling an RL algorithm for concrete decision-making tasks.

\bibliography{literature}

\end{document}